\documentclass{article} 
\usepackage{iclr2024_conference,times}

\usepackage{amsmath,amsfonts,bm}

\def\eqref#1{equation~\ref{#1}}

\def\1{\bm{1}}

\def\vd{{\bm{d}}}

\def\vp{{\bm{p}}}

\def\mA{{\bm{A}}}

\def\mE{{\bm{E}}}

\def\mI{{\bm{I}}}

\def\mM{{\bm{M}}}

\def\mP{{\bm{P}}}
\def\mQ{{\bm{Q}}}

\def\mW{{\bm{W}}}
\def\mX{{\bm{X}}}
\def\mY{{\bm{Y}}}

\DeclareMathAlphabet{\mathsfit}{\encodingdefault}{\sfdefault}{m}{sl}
\SetMathAlphabet{\mathsfit}{bold}{\encodingdefault}{\sfdefault}{bx}{n}

\usepackage[colorlinks, pagebackref=true, citecolor=blue, urlcolor=blue, linkcolor=blue]{hyperref}
\usepackage{url}
\usepackage{float}
\usepackage{enumitem}
\usepackage{graphicx}
\usepackage{booktabs}
\usepackage{setspace}
\usepackage{amsthm}
\usepackage{adjustbox}
\usepackage{multirow}
\usepackage[skip=3pt]{caption}
\usepackage{dsfont}
\usepackage{subcaption}

\definecolor{lightgray}{gray}{0.97}

\usepackage{color, colortbl}

\usepackage{wrapfig}

\usepackage{algorithm}
\usepackage{algorithmicx}
\usepackage[noend]{algpseudocode}

\theoremstyle{plain}
\newtheorem{theorem}{Theorem}[section]
\newtheorem{proposition}[theorem]{Proposition}
\newtheorem{lemma}[theorem]{Lemma}

\theoremstyle{definition}

\theoremstyle{remark}

\usepackage{xcolor}
\definecolor{Gray}{gray}{0.93}

\definecolor{myforestgreen}{RGB}{70, 160, 70}

\renewcommand{\eqref}[1]{Eqn.~(\ref{#1})}

\newcommand{\spmb}[1]{\fontsize{4}{6}\selectfont{{\textcolor{blue}{\pm#1}}}}

\title{Sparse Training of Discrete Diffusion Models for Graph Generation}

\author{Yiming Qin,\  Clément Vignac,\  Pascal Frossard \thanks{Contact: \texttt{first\_name.last\_name@epfl.ch}}\\ LTS4, EPFL, Lausanne, Switzerland}

\begin{document}

\maketitle

\begin{abstract}
Generative graph models struggle to scale due to the need to predict the existence or type of edges between all node pairs. To address the resulting quadratic complexity, existing scalable models often impose restrictive assumptions such as a cluster structure within graphs, thus limiting their applicability.
To address this, we introduce SparseDiff, a novel diffusion model based on the observation that almost all large graphs are sparse. By selecting a subset of edges, SparseDiff effectively leverages sparse graph representations both during the noising process and within the denoising network, which ensures that space complexity scales linearly with the number of chosen edges. During inference, SparseDiff progressively fills the adjacency matrix with the selected subsets of edges, mirroring the training process.
Our model demonstrates state-of-the-art performance across multiple metrics on both small and large datasets, confirming its effectiveness and robustness across varying graph sizes. It also ensures faster convergence, particularly on larger graphs, achieving a fourfold speedup on the large Ego dataset compared to dense models, thereby paving the way for broader applications.
\footnote{Our code is available at \url{https://github.com/qym7/SparseDiff}.}
\end{abstract}

\section{Introduction}\label{sec:introduction}

Over the past decades, random graph models have played a foundational role in graph generation \citep{erdHos1960evolution, barabasi2013network}.
However, recent advances in capturing complex data distributions have redirected research efforts towards learned graph models.
In this domain, traditional frameworks such as generative adversarial networks \citep{de2018molgan} and variational autoencoders \citep{simonovsky2018graphvae}, which were primarily focused on small graphs, have been succeeded by denoising diffusion models \citep{niu2020permutation, jo2022score, vignac2022digress}. These diffusion models not only set new benchmarks in various graph generation tasks but also partially addressed the scalability limitations, capable of generating graphs with up to 200 nodes.
However, current diffusion models still struggle to scale to large graphs required for broader applications such as transportation \citep{rong2023city} or anomaly detection in financial systems \citep{li2023diga}.

The primary complexity constraint in graph generative models arises from the need to predict interactions for all node pairs, leading to quadratic computational complexity relative to the number of nodes. Methods to mitigate this limitation include imposing node ordering \citep{dai2020scalable}, assembling sub-graphs \citep{limnios2023sagess}, hierarchical generation \citep{karami2023higen, jang2023hggt, bergmeister2023efficient}, and conditioning on a sampled degree distribution \citep{chen2023efficient}. While these approaches are designed for large graphs, they require additional assumptions such as cluster structures or dependencies on node degrees. Their suboptimal performance in applications like molecule generation \citep{chen2023efficient, kong2023autoregressive} casts doubt on the robustness of these assumptions.

Motivated by the above limitations, and considering that real-world graphs are mostly sparse, we propose SparseDiff, a discrete diffusion model that only harnesses the inherent sparsity of training graphs without imposing extra assumptions, enabling more general application to diverse downstream tasks.
Specifically, SparseDiff employs a sparse representation based on edge lists to address the quadratic space complexity inherent in three key components of a diffusion model. It introduces: 1) A computationally efficient noise model that maintains the sparsity of noisy graphs, 2) A sparse denoising network trained on a random subset of node pairs, 3) An iterative sampling procedure designed to progressively populate the dense adjacency matrix with sparse inputs. Each component maintains linear complexity with respect to the number of edges while enabling it to handle graphs of varying sizes effectively.
By specifying the size of the subset of edges used, SparseDiff ensures controllable space complexity during both the training and inference phases. Our work represents the first application of sparse representations for graph diffusion, significantly differentiating itself from previous methods based on hierarchical or autoregressive frameworks.

Our experiments demonstrate that SparseDiff not only achieves state-of-the-art performance but also converges significantly faster than dense diffusion models on large graphs such as protein interaction networks (Protein) and social networks (Ego). Remarkably, it also outperforms dense models like SPECTRE \citep{martinkus2022spectre} and DiGress\citep{vignac2022digress}, as well as other scalable models like EDGE \citep{chen2023efficient} and HGGT\citep{jang2023hggt}, in managing smaller molecular datasets like QM9 and Moses. These results empirically validate the effectiveness of our sparse modeling approach and confirm its robustness across a diverse range of datasets without relying on additional assumptions.

\section{Related Work}

\subsection{Denoising Diffusion Models for Graphs}
Diffusion models have gained increasing popularity due to their impressive performance across various generative tasks in fields such as computer vision \citep{dhariwal2021diffusion, ho2022imagen}, protein generation \citep{baek2021accurate, ingraham2022illuminating}, and audio synthesis \citep{kong2020diffwave}. These models are characterized by three core components. The first is a Markovian noise model, which progressively corrupts a data point $x$ to a noisy sample $z^t$ over iterative steps $t$ from $1$ to $T$, until it conforms to a predefined prior distribution at step $T$. The second component is a denoising network, parametrized by $\theta$, which is trained to restore the corrupted data back to its less noisy state. Typically, this network aims to predict the original data $x$ given the noisy sample $x_t$. The third component is the reverse process for data generation, where a fully noisy data point $z^T$ is first sampled from a prior distribution. The denoising network then operates at each time step $t \in [T, ..., 1]$ to predict the less noisy distribution according to $p_\theta(z^{t-1} | z^t) = \int_x q(z^{t-1} | z^t, x) \, dp_\theta(x|z^t)$, from which a new data point $z^{t-1}$ is sampled. While this integral is generally difficult to evaluate, two prominent frameworks, Gaussian diffusion \citep{ho2020denoising} and discrete diffusion \citep{austin2021structured}, facilitate its efficient computation.

Initial graph diffusion models employed Gaussian noise directly on adjacency matrices \citep{niu2020permutation, jo2022score}. These models use a graph attention network to regress the added noise $\epsilon$, where $\epsilon = z^t - z$, effectively regressing the noise up to an affine transformation, akin to regressing the discrete clean graph. To preserve the inherent discreteness of graphical data, subsequent models \citep{vignac2022digress, haefeli2022diffusion} have leveraged discrete diffusion, reformulating graph generation as a series of classification tasks, and achieving top-tier results. However, these models require predictions for all pairs of nodes, which implies a quadratic space complexity and thus restricts their scalability.

\subsection{Scalable Graph Generation}

Efforts to enhance the scalability of graph generative models mainly follow two paradigms: hierarchical refinement and subgraph aggregation.
The hierarchical refinement approach initially generates a low-resolution graph, which undergoes successive refinements for enhanced detail \citep{yang2020scaleauto,karami2023higen}. For instance, the HGGT model \citep{jang2023hggt} employs a hierarchical $K^2$-tree representation. For molecule generation, fragment-based models \citep{jin2018junction, jin2020hierarchical, maziarz2021learning} adeptly assemble compounds using pre-defined molecular fragments. Recently, \citet{bergmeister2023efficient} proposes a hierarchical diffusion model based on spectrum-preserving local expansion algorithms, enabling the generation of non-attributed large graphs.
On the other side, the subgraph aggregation approach divides larger graphs into smaller subgraphs, which are subsequently combined. For instance, SnapButton \citep{yang2020scaleauto} enhances autoregressive models \citep{liu2018constrained, liao2019efficient, mercado2021graph} by merging subgraphs, and SaGess \citep{limnios2023sagess} trains a dense diffusion model to generate subgraphs sampled from a large graph that are then merged.

Additionally, some approaches predict all node pairs in an auto-regressive way. \citet{kong2023autoregressive} integrated diffusion with autoregressive models, suggesting learning the node ordering, which is theoretically as difficult as isomorphism testing. On the other hand, EDGE \citep{chen2023efficient} uses absorbing states to create sparse diffusion. It starts by generating a node degree distribution $\vd^0$ and gradually constructs an adjacency matrix $\mA$ based on the changes of node degrees during inference. While this factorization is universally applicable, the feasibility of learning the conditional distribution $p_\theta(\mA | \vd^0)$ remains uncertain, as not all degree distributions can be achieved by undirected graphs.

Overall, scalable generation models typically either introduce a dependence on node orderings or rely heavily on extra assumptions on data distribution. In contrast, the SparseDiff model described in the next section aims at making no assumption besides graph sparsity, while showing competitive performance compared to other scalable models across a wide range of graphs.

\begin{figure*}[t]
\begin{center}
\includegraphics[width=0.99\textwidth]{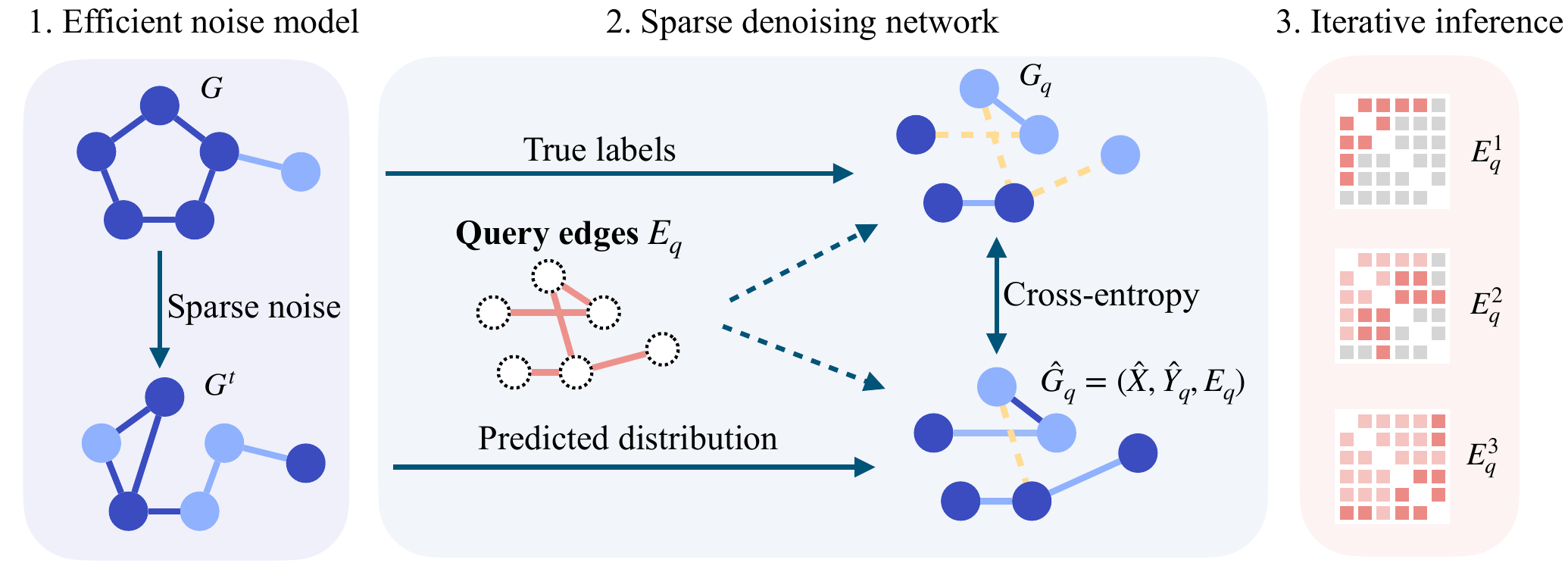}
\end{center}
\caption{
SparseDiff has three main components. First, to define the diffusion process, it constructs a sparse noise model that samples noise trajectories with linear space complexity relative to the number of edges.
During network training, rather than considering all node pairs, SparseDiff randomly selects a subset of "query edges" ($\mE_q$) based on a sparsity parameter $\lambda$, set to $\frac{1}{3}$ for this visualization. $\mE_q$ includes both 'existing' edges from the original graph and 'non-existing' edges, which will potentially become new edges, enabling the model to learn all edge types while efficiently managing space complexity.
In the inference phase, SparseDiff predicts labels only for the query edges $\mE_q^i$ in each iteration $i$, ensuring consistency with the training phase.
}
\label{fig:forward}
\end{figure*}

\section{SparseDiff: Sparse Discrete Diffusion for Graph Generation}\label{sec:our_approach}

We now introduce SparseDiff, a Sparse Denoising Diffusion Model designed to enhance the scalability of graph diffusion and extend the reach of discrete graph models to graphs with up to $1,000$ nodes, a significant improvement over previous dense models limited to around $200$ nodes.

Unlike previous graph diffusion models, SparseDiff builds on a sparse representation of graphs. A graph $G$, consisting of $n$ nodes and $m$ edges, is represented as a triplet $(\mE, \mX, \mY)$. Here, $\mE \in \mathbb{N}^{2\times m}$ indicates the edge list with the indices of endpoints. Node and edge attributes are considered to be discrete and are encoded in a one-hot format as $\mX \in \{0, 1\}^{n\times a}$ and $\mY \in \{0, 1\}^{m\times b}$, where $a$ and $b$ are the number of classes, respectively. In particular, `non-existing' edges are considered an additional edge type, while edges in $\mE$ are referred to as `existing' edges. This sparse representation is widely supported by standard graph processing packages such as Pytorch Geometric \citep{fey2019fast}.

This work focuses solely on undirected graphs with discrete attributes, although continuous labels can be seamlessly integrated, as demonstrated by \citet{Vignac2023MiDiMG}. All considered graphs are free of self-loops.
Figure \ref{fig:forward} provides an overview of the SparseDiff framework. Further details on the training and sampling processes are provided in Algorithms \ref{alg:train} and \ref{alg:sample}, respectively. In the following parts, we specifically focus on three critical components of the proposed sparse diffusion model for graphs: the noise model, the denoising network, and the iterative sampling algorithm.

\subsection{Efficient Noise Model}
\label{subsec:noise_model}
To improve memory efficiency, SparseDiff employs a dedicated noise model that maintains a similar sparsity level of the noisy graph $G^t$ throughout the noising process. This model also reduces the space complexity of computing noisy graphs from $O(n^2)$ to $O(m)$.

\subsubsection{Sparse Trajectory}
Given that Gaussian noise applied to the adjacency matrix typically results in dense noisy graphs, where all edge entries acquire continuous values \citep{niu2020permutation, jo2022score}—we opt for a discrete diffusion framework. In this framework, we only need to consider the existing edges within a noisy graph, thereby reducing the number of edges necessary for computation.

In the discrete graph diffusion model, the noisy trajectory at each step is defined by $q(G^t|G^{t-1}) = (\mX \mQ^t_\mX, \mY \mQ^t_\mY)$, where $\mQ^t$ denotes the Markov transition matrix for that step, which transforms $G^{t-1}$ into a noisier distribution. Various types of Markov transition matrices are employed, including uniform, absorbing \citep{austin2021structured}, and marginal transitions \citep{vignac2022digress}. We employ the marginal transition model, which favors transitions towards the dominant class in the marginal distribution. This strategy is particularly effective for preserving sparse trajectories in graphs by naturally biasing the model toward `non-existing' edges. Formally, considering the marginal distribution vectors $\vp_\mX$ for node types and $\vp_\mE$ for edge types, and denoting their transposes by $\vp'$, the noise level at each step $t$ is regulated by $\beta^t$, with $\alpha^t = 1 - \beta^t$. The marginal transition matrices are defined as follows:
\[
\mQ^t_{\mX} = \alpha^t \mI + \beta^t \mathbf{1}_a \vp'_\mX, \quad \mQ^t_{\mY} = \alpha^t \mI + \beta^t \mathbf{1}_b \vp'_\mY.
\]
Here, \(\mathbf{1}_a\) and \(\mathbf{1}_b\) are column vectors of ones with dimensions equal to the number of classes \(a\) for nodes and \(b\) for edges. These matrices incorporate a first term, the identity matrix \(\mI\), to preserve the distribution from \(G^{t-1}\), and a second term to introduce noise aligned with the marginal distributions.

By employing continuous multiplication, we can derive the distribution at step $t$ directly from the initial clean graph using $q(G^t|G) = (\mX \bar\mQ^t_\mX, \mY \bar\mQ^t_{\mY})$, facilitating an immediate transition to the noisy state without the need for iterative step-by-step calculations. For instance, the cumulative transition matrix $\bar{\mQ}^t$ for nodes $\mX$ is as follows: $\bar{\mQ}^t_\mX = \mQ^1_\mX \mQ^2_\mX \ldots \mQ^t_\mX = \bar\alpha^t \mI + (1-\bar\alpha^t) \mathbf{1}_a \vp'_\mX$, where $\bar\alpha^t = \alpha^1 \alpha^2 \ldots \alpha^t$. The parameter $\bar\alpha^t$ starts very close to $1$ at $\bar\alpha^1$ and approaches $0$ by $\bar\alpha^T$, reflecting a gradual increase in noise influence over diffusion process.

We note that this choice of marginal noise model does not guarantee that the noisy graph is always sparse. However, it is the case with high probability, as stated by the following lemma, which is an application of \citet{desolneux2008estimating} (detailed in Appendix \ref{app:math}).

\vspace{5pt}
\begin{lemma} (Probability Bound for Sparsity in Noisy Graphs) \label{thm:tail}
Consider an undirected graph with $n$ nodes, $m$ edges, and no self-loops. If the edge ratio given by $m / \left(\frac{n(n-1)}{2}\right)$ is denoted as $r$, and the edge ratio in the noisy graph sampled from the marginal transition noise model is given by $r_t$, then for $n$ sufficiently large and $r<\frac{1}{4}$, for any $r < k < 1$, we have:
\begin{equation}
    \log (\mathrm{P}[r_t \geq k]) \sim -\frac{n(n-1)}{2} \left(k \log \frac{k}{r} + (1-r) \log \frac{1-k}{1-r}\right)
\end{equation}
\end{lemma}

This lemma shows that, in large and sparse graphs, the probability of the edge ratio $r_t$ in the noisy graph exceeding $k$, where $k > r$, declines exponentially with the graph size. As an example, for $r$ small and $k = 2r$, this probability is approximated by $c_1 e^{-c_2 n^2 r}$, where $c_1$ and $c_2$ are constants. This probability diminishes significantly as the graph size $n$ increases.

\subsubsection{Sparse Computation}

\begin{figure*}[t]
\begin{center}
\includegraphics[width=0.99\textwidth]{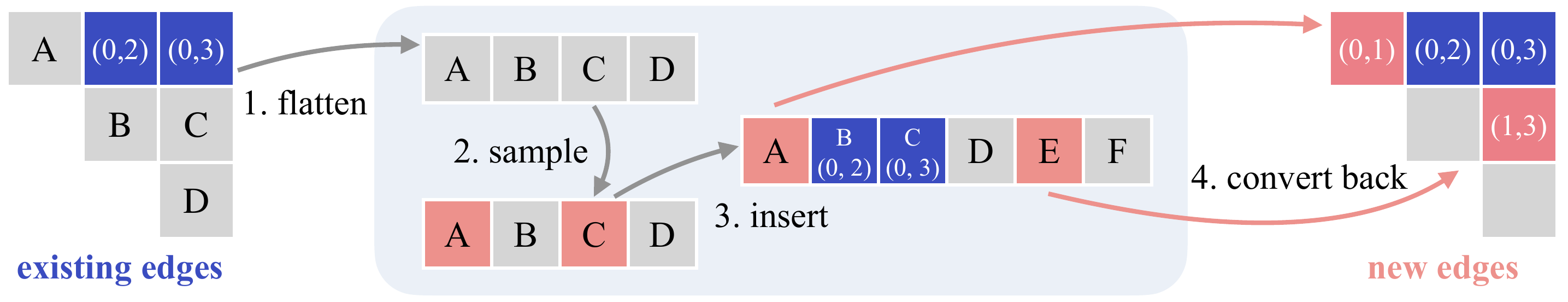}
\end{center}
\caption{
Efficient sampling of new edge positions among `non-existing' positions using only the edge list, thus avoiding the quadratic space complexity of an adjacency matrix. The process involves: 1) \textbf{Flattening} the pair representations into a linear array, 2) \textbf{Sampling} uniformly from four non-occupied positions, selecting A (1$^{st}$) and C (3$^{rd}$), 3) \textbf{Inserting} offsets of 0 and 2 to the positions of A and C to account for the count of `existing' edges before the selected positions, resulting in positions A (1$^{st}$) and E (5$^{th}$), 4) \textbf{Converting} these positions back to index pairs $(0, 1)$ and $(1, 3)$.
}
\label{fig:sample_algo}
\end{figure*}

Our second requirement for the noise model is subquadratic space complexity sampling. Standard discrete diffusion models encode all edges using $\mY \in \mathbb{R}^{n \times n \times b}$ and calculate transition probabilities with $\mY \bar{\mQ}^t_{\mY} \in \mathbb{R}^{n \times n \times b}$, which incurs $O(n^2)$ space complexity. To facilitate sparse sampling, we differentiate between `existing' and `non-existing' edges. In sparse graphs, where `existing' edges are a minority, we compute $\mY \bar{\mQ}^t_{\mY} \in \mathbb{R}^{m \times b}$ for these edges and sample their new labels from this distribution. Edges transitioning to `non-existing' are then removed to preserve sparsity in $G^t$.
Conversely, for the majority `non-existing' edges in sparse graphs, we use a novel three-step approach to sample without using dense adjacency matrices. The process involves:

\begin{enumerate}
    \item Sampling the number of new `existing' edges.
    \item Sampling positions for new edges uniformly from non-occupied node pairs (c.f. Fig. \ref{fig:sample_algo}).
    \item Sampling the edge attributes from all labels except `non-existing'.
\end{enumerate}

Specifically, the number of new edges, that is, `non-existing' edges transitioning to `existing', follows a Binomial distribution $\mathcal{B}(\bar{m}_t, q_t)$. Here, $\bar{m}_t = \frac{n(n-1)}{2} - m_t$ represents the total possible edges minus the `existing' ones in $G^t$. With `non-existing' as the first class, $q_t = 1 - \mQ^t[0,0]$ indicates the transition probability from `non-existing' to `existing' edge types. New edge labels are determined by a Multinomial distribution as specified by $\mQ^t[0,1:\ ]$. This method markedly improves efficiency and scalability in modeling large graphs.

The second step, sampling new edge positions, presents more challenges. As illustrated in Figure \ref{fig:sample_algo}, our algorithm efficiently handles the sparse sampling without requiring a full adjacency matrix. The method begins by flattening the graph's vacant position pair representations into a linear array. We then perform simple random sampling from this array and then adjust the sampled indices by adding offsets to accommodate existing edges, before remapping them to index pairs. This process accommodates simultaneous sampling across batches of graphs with varying sizes, ensuring efficient space usage and precise placement of new edges.

\subsection{Efficient Denoising Neural Network}
\label{subsec:arch}

Unlike traditional graph diffusion models that encode features for every node pair with a computational complexity of $O(l n^2 d_e)$, where $l$ represents the layer count and $d_e$ the dimension of edge activations, our approach builds on a memory-efficient graph neural network that utilizes edge lists, significantly reducing the complexity to $O(l m d_e)$, where $m$ is the number of chosen edges involved in message-passing. This efficiency is achieved by leveraging sparse representations within message-passing neural networks (MPNNs) \citep{scarselli2008graph, gilmer2017neural}, utilizing specialized libraries like Pytorch Geometric \citep{fey2019fast} or the Deep Graph Library \citep{wang2019deep}.

Specifically, our architecture differs from most message-passing networks by needing to make predictions for all node pairs instead of `existing' ones.
This shift explains why previous models employed graph transformer networks \citep{vignac2022digress}, which inherently encode each node pair within their full attention layers. To circumvent this need, a common strategy for knowledge graphs \citep{zhang2018link, chamberlain2022graph, boschin2023machine} involves predicting edge labels based on the pairwise similarity between node embeddings. Although theoretically feasible, this link-prediction method fails to reach the performance of dense denoising models (detailed in Table \ref{tab:link} in Appendix \ref{app:ablations}), even on small graph datasets. Consequently, edges for prediction have to be explicitly encoded to enhance prediction accuracy in practice.

\subsubsection{Edge Prediction Module}
\label{subsub:edgeemb}

\begin{figure*}[t]
\includegraphics[width=0.99\textwidth]{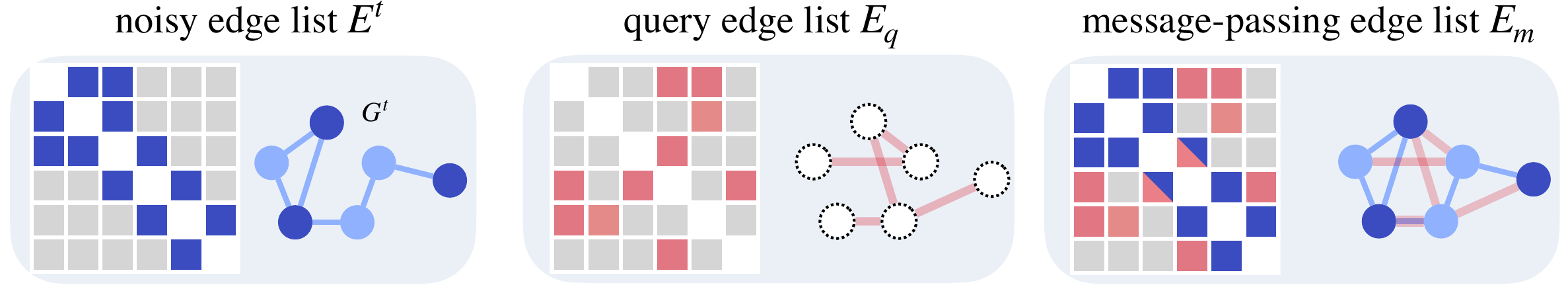}
\centering
\caption{
The message-passing edge list $\mE_m$, fed to the sparse denoising network, comprises noisy edges $\mE^t$ (shown in blue) and uniformly sampled query edges $\mE_q$ (shown in red).
}
\label{fig:query}
\end{figure*}

As previously discussed, relying solely on node features for edge prediction yields limited results. To enhance edge prediction, we randomly select a subset of node pairs for explicit encoding during each forward pass, termed `query edges'. As indicated in Figure \ref{fig:query}, the input to the denoising network, referred to as message-passing edges $(\mE_m, \mY_m)$, thus includes two sets of edges: `noisy edges' $(\mE^t, \mY^t)$, representing `existing' edges in the noisy graph $G^t$, and `query edges' $(\mE_q, \mY_q)$, which are random edges selected for prediction. Noisy edges preserve the topological information of the noisy graph, while query edges, sampled randomly from all node pairs, include both `existing' and `non-existing' edges, enabling the model to predict all edge types. To provide an unbiased estimator of the loss relative to dense diffusion models, query edges are uniformly sampled across each graph.

Additionally, the `non-existing' edges included in $\mE_q$ serve as shortcuts within the message-passing network, a mechanism that facilitates graph rewiring. This is known to enhance information propagation and mitigate over-squashing issues, as noted in studies like \citep{alon2020bottleneck, topping2021understanding, pmlr-v202-di-giovanni23a}.

To control the number of query edges, we define the `sparsity parameter' $\lambda$ as the ratio of query edges to all node pairs. Given that query edges may overlap with noisy edges and the number of noisy edges $m_t$ approximates the number of edges $m$ in the clean graph according to Lemma \ref{thm:tail}, the total number of message-passing edges is approximately upper bounded by $(m + \lambda n^2)$. By setting $\lambda = \frac{m}{n^2}$, the computational complexity of SparseDiff practically aligns linearly with the number of edges $m$ in the clean graph $G$.
In our experiments, $\lambda$ is chosen to enable an effective batch size for training, as detailed in Appendix \ref{app:stats}.
This parameter provides flexibility under varying computational conditions.

\subsubsection{Model Training}

Our sparse denoising network adopts a graph transformer architecture featuring normalization, feed-forward, and attention layers \citep{velivckovic2017graph}. It incorporates a sparse attention mechanism for handling sparse data \citep{shi2020masked}, and integrates advanced features such as PNA pooling layers \citep{corso2020principal} and FiLM layers \citep{perez2018film}, which are designed to enhance predictive accuracy and effectively manage computational complexity. A detailed discussion of the model architecture is provided in Appendix \ref{app:arch}.

Training of the network involves predicting `query edges' $\mE_q$, and the loss is minimized using the cross-entropy (CE) loss between the predicted distribution $\hat{\mP}^G_q = (\hat{\mP}^X, \hat{\mP}^Y_q)$ and the clean graph $G$. The loss function is computed as follows:
\begin{align}
\sum\limits_{1 \leq i \leq n} \operatorname{CE}(\mX_i, \hat{\mP}^\mX_i) + \frac{c}{\lambda} \sum\limits_{(i,j)\in \mE_q} \operatorname{CE}(\mY_{ij}, \hat{\mP}^\mY_{ij}), \label{eq:loss}
\end{align}
Here, the constant \(c\) weights nodes and edges in the loss calculation. It is rescaled by dividing by \( \lambda \) to maintain a consistent edge-to-node weight ratio across different \( \lambda \) values.

\begin{figure}
  \begin{minipage}[t]{1.0\linewidth}
    \begin{algorithm}[H]
      \caption{Sparse training at step $t$ with the sparsity parameter $\lambda$ (Section \ref{subsec:noise_model} $\&$ \ref{subsec:arch})}
      \begin{algorithmic}[1]
        \State Given the clean graph $G=(\mE^0, \mX^0, \mY^0)$;
        \State Sample the noisy graph $G^t = (\mE^t, \mX^t, \mY^t)$;
        \State Sample query edges $\mE_q$ of size $\lceil\lambda n^2\rceil$ ;
        \State $\mE_m \leftarrow \mE^t \cup \mE_q$, $\mY_m \leftarrow \mY^t \cup \mY_q$;\Comment{Construct message-passing edges}
        \State $G_m\leftarrow(\mE_m, \mX^t, \mY_m)$;\Comment{Construct the message-passing graph}
        \State $(\hat{\mP}^\mX, \hat{\mP}^\mY_q) = \phi_\theta(G_m, \mE_q)$;\Comment{Predict the distribution of nodes and query edges}
        \State $\operatorname{optimizer.step}(\operatorname{CE}(\hat{\mX^0},  \hat{\mP}^\mX) + \operatorname{CE}(\mY^0_q, \hat{\mP}^\mY_q)$);\Comment{Loss calculation}
      \end{algorithmic}
      \label{alg:train}
    \end{algorithm}
  \end{minipage}
  \hfill
\end{figure}
\hfill
\begin{figure}
  \begin{minipage}[t]{1.0\linewidth}
    \begin{algorithm}[H]
      \caption{Iterative inference at step $t$ with the sparsity parameter $\lambda$ (Section \ref{subsuc:sample})}
      \begin{algorithmic}[1]
        \State Initialize an empty graph $G^{t-1}$ with unlabeled nodes $\mX^{t-1}$ and no edges;
        \State Randomly divide all node pairs into $K = \lceil \frac{1}{\lambda} \rceil$ equal-sized chunks $\{C_0, \cdots, C_{K-1}\}$;
        \For{k in range($K$)}
            \State $\mE_q \leftarrow C_{k}$;\Comment{Set query edges}
            \State $\mE_m \leftarrow \mE^t \cup \mE_q$;\Comment{Construct message-passing edges and its attributes $\mY_m$}
            \State $G_m\leftarrow(\mE_m, \mX^t, \mY_m)$;\Comment{Construct the message-passing graph}
            \State ($\hat{\mP^\mX}$, $\hat{\mP^\mY_q}) = \phi_\theta(G_m, \mE_q)$;\Comment{Predict the distribution of nodes and query edges}
            \State $\hat{\mX}=\operatorname{Multinomial}(\hat{\mP}^\mX), \hat{\mY_q}=\operatorname{Multinomial}(\hat{\mP}^\mY_q)$;\Comment{Sample labels}
            \State $\mX^{t-1} \leftarrow \hat{\mX}$;\Comment{Assign node new labels}
            \State $\mY^{t-1}\leftarrow\mY^{t-1}\cup\hat{\mY_q}[\hat{\mY_q}!=0], \mE^{t-1}\leftarrow\mE^{t-1}\cup\hat{\mE_q}[\hat{\mY_q}!=0]$;\Comment{Add `existing' edges}
        \EndFor
      \end{algorithmic}
      \label{alg:sample}
    \end{algorithm}
  \end{minipage}
  \hfill
\end{figure}

\subsection{Iterative Inference}
\label{subsuc:sample}

SparseDiff also remains memory-efficient at the inference stage. Here, we first sample the number of nodes $n$ from the node distribution of the training set, which remains constant during the reverse process. Next, we sample a random graph from the prior distribution $G^T \sim \prod_{i=1}^n \operatorname{Cate}(\vp_\mX) \times \prod_{1 \leq i < j \leq n} \operatorname{Cate}(\vp_\mY)$, where $\vp_\mX$ and $\vp_\mY$ represent the marginal probabilities of node and edge classes, respectively. The categorical distribution $\operatorname{Cate}(\vp)$ is used for both nodes and edges. The sparse denoising network $\phi_\theta$ is then recursively applied to predict the clean graph from the noisy one. The generation process of SparseDiff is visualized in Figure \ref{fig:iter_sampling}.

Directly predicting the entire graph at each diffusion step $t$ is impractical due to quadratic memory requirements. Moreover, using dense graphs during inference could lead to a distribution shift caused by changes in the number of edges used for message-passing. To mitigate this, we implement an iterative procedure to progressively cover all node pairs in $G^{t-1}$. As detailed in Algorithm \ref{alg:sample}, we divide all node pairs randomly into $K=\lceil \frac{1}{\lambda} \rceil$ equally-sized sets, representing the query edges for each prediction step\footnote{When $\frac{n(n-1)}{2}$ is not divisible by $K$, we adjust by slightly overlapping the last set with the previous one.}. During each iteration $k$, the noisy graph $G^t$ remains identical, and a message-passing edge list $\mE_m$ is constructed using noisy edges and query edges from the $k^{th}$ set. SparseDiff then predicts the distributions for these query edges, samples labels, and integrates edges classified as 'existing' into $G^{t-1}$.

This iterative approach allows SparseDiff to maintain favorable memory complexity, albeit at the cost of increased sampling time due to iterations at each diffusion step. However, unlike many scalable models, SparseDiff does not impose additional assumptions about data distribution, such as clustering or degree distribution. Despite the increased sampling time, the generation time for SparseDiff remains efficient due to its ability to utilize larger batch sizes and to accelerate model computations using sparse inputs, as reported in Table \ref{tab:sampling_speed} of Appendix \ref{app:sampling_speed}.

Besides, drawing inspiration from D3PM \citep{austin2021structured} and DDIM \citep{song2020denoising}, we propose a method to accelerate inference by reducing the inference steps by a factor $k$. 
In particular, at each step $t$, the model predicts $q(G^{t-k}|G^t,G) \propto G^t(Q^t)'\odot G\bar{Q}^{t-k}$ instead of $q(G^{t-1}|G^t,G) \propto G^t(Q^t)'\odot G\bar{Q}^{t-1}$. The results for acceleration are reported in Table \ref{tab:acceleration_results}.

\begin{figure*}[t]
    \centering
    \begin{subfigure}[b]{0.49\linewidth}
         \centering
         \includegraphics[width=0.99\linewidth]{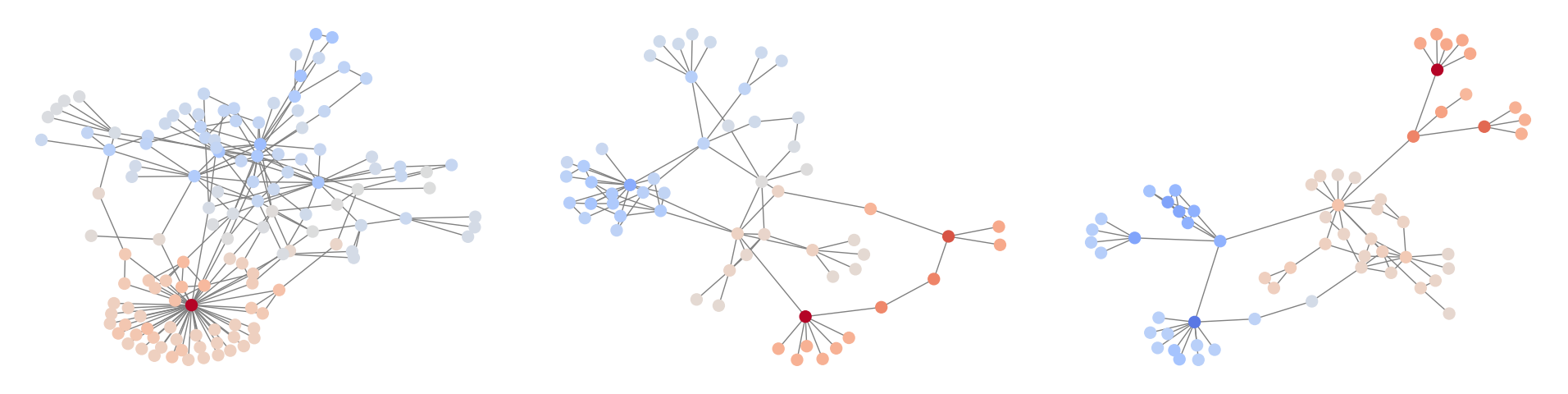}
         \caption{}
    \end{subfigure}
    \hfill
    \begin{subfigure}[b]{0.49\linewidth}
         \centering
         \includegraphics[width=0.99\linewidth]{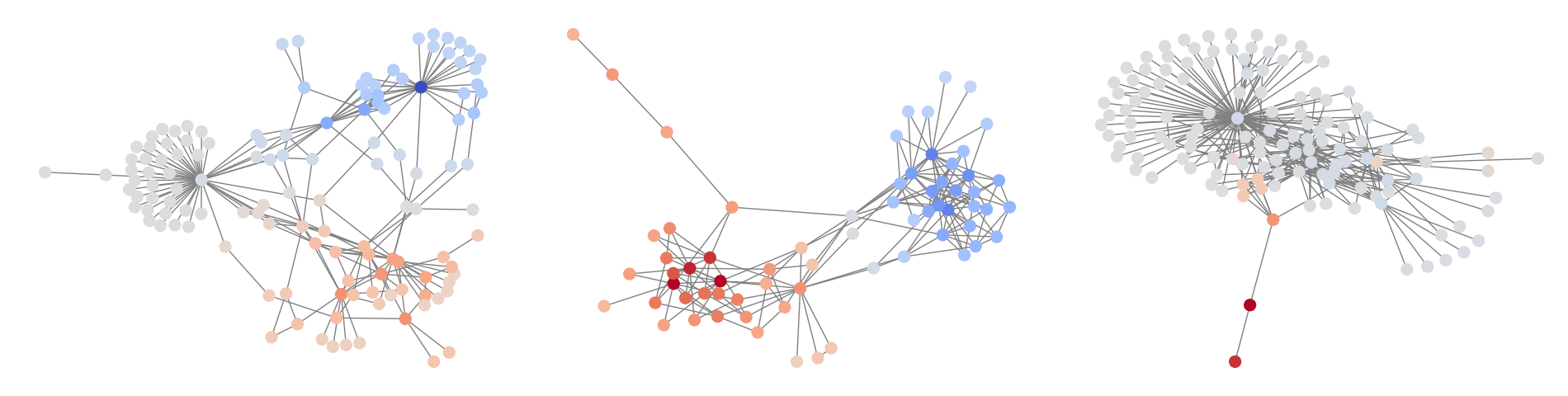}
         \caption{}
    \end{subfigure}
    \hfill
    \begin{subfigure}[b]{0.49\linewidth}
         \centering
         \includegraphics[width=0.99\linewidth]{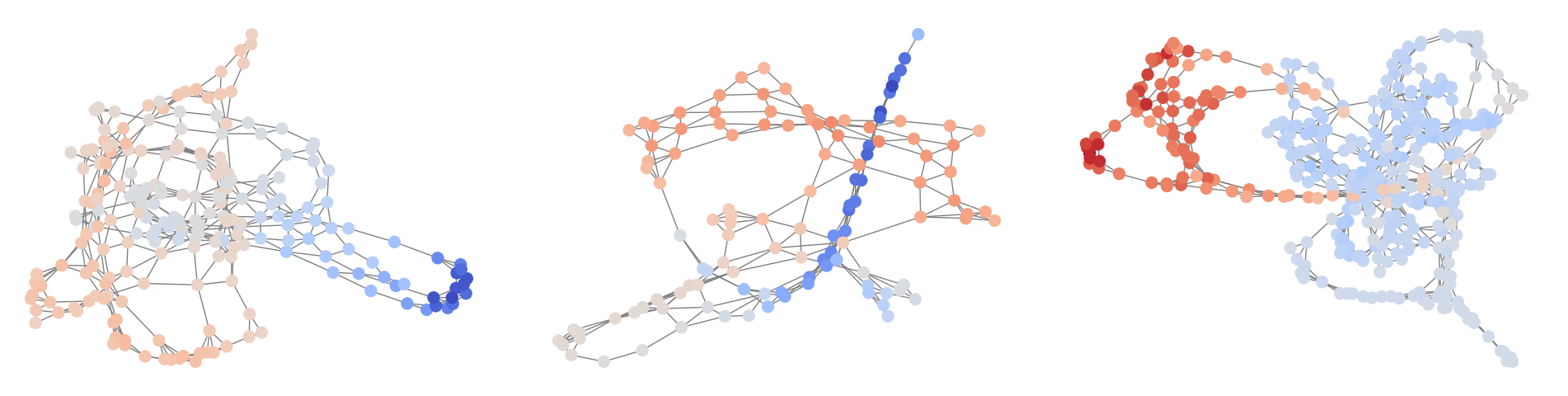}
         \caption{}
    \end{subfigure}
    \begin{subfigure}[b]{0.49\linewidth}
         \centering
         \includegraphics[width=0.99\linewidth]{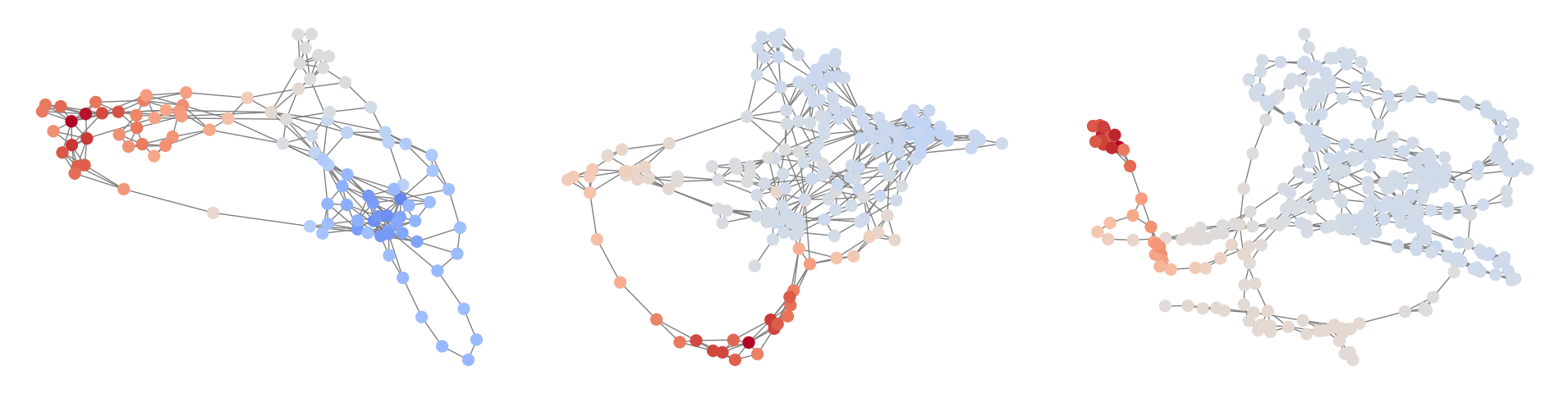}
         \caption{}
    \end{subfigure}

    \caption{Samples from SparseDiff trained on large graphs. (a) Ego training set ($50$ to $399$ nodes); (b) Generated Ego graphs; (c) Protein training set ($100$ to $500$ nodes); (d) Generated Protein graphs.
    }
    \label{fig:vis_large}
\end{figure*}

\section{Experiments}
\label{sec:exp}

We conducted comprehensive experiments to evaluate the performance of SparseDiff across a diverse set of graphs comparing against a comprehensive range of models. These include GraphRNN \citep{you2018graphrnn}, GRAN \citep{liao2019efficient}, GraphNVP \citep{madhawa2019graphnvp}, SPECTRE \citep{martinkus2022spectre}, GDSS \citep{jo2022score}, DiGress \citep{vignac2022digress}, DruM \citep{Jo2023GraphGW}, and several scalable models such as BiGG \citep{dai2020scalable}, GraphARM \citep{kong2023autoregressive}, EDGE \citep{chen2023efficient}, HiGen \citep{karami2023higen}, and HGGT \citep{jang2023hggt}. For clarity, we refer to the local expansion method proposed by \citet{bergmeister2023efficient} as GraphLE.

To ensure reliability, SparseDiff's performance metrics are presented in the format of mean $\pm$ standard deviation, based on five samplings. This accounts for the instability caused by limited test set sizes in some datasets, namely SBM and Planar datasets. We bold SparseDiff's results if the best results fall within our method's variance range, indicating that with an optimal seed, SparseDiff can achieve the best performance.
The results highlight SparseDiff's strong competitive advantage on datasets containing larger graphs, such as Planar, SBM, Protein, and Ego, as well as its state-of-the-art performance on datasets with small molecules, including QM9 and Moses (as detailed in Appendix \ref{app:moses}). Notably, SparseDiff is the only model thus far that exhibits competitive performance across both large and small graphs, and in both scenarios of attributed and unattributed graphs.

\subsection{Large Graph Generation}
\label{subsec:large}

\paragraph{Dataset}
We evaluate SparseDiff on diverse graph datasets to demonstrate its scalability and versatility. First, we test its ability to generate edge-crossing-free planar graphs with 64 nodes. Next, we assess its capacity to generate graphs with 2 to 5 communities using Stochastic Block Model (SBM) graphs, scaling up to 200 nodes—the largest size seen in models like DiGress \citep{vignac2022digress}. We also evaluate Ego and Protein datasets, with graphs up to 500 nodes, representing citation relationships and amino acid interactions within 6 Angstroms. The largest edge ratio for these datasets is $8.8\%$, confirming their sparsity. Detailed dataset statistics are in Appendix \ref{app:stats}. To further highlight our model's scalability, we include the generation of a large graph with 1,045 nodes in Appendix \ref{app:single_graph}.

\paragraph{Metrics}
For evaluation, we use maximum mean discrepancy (MMD) metrics, standard in graph generation tasks. We report the validity of SBM graphs as the fraction passing a stochastic block model test, and for Planar graphs, the fraction that are planar and connected. For larger datasets, we also use the Radial Basis Function (RBF) MMD metric to assess fidelity and diversity using a randomly parametrized GNN \citep{thompson2022evaluation}.
Since MMD metrics often yield small values that are difficult to compare directly, we report Degree, Cluster, Orbit, Spectre and RBF MMD metrics in units of $10^{-3}$, $10^{-2}$, $10^{-2}$, $10^{-3}$ and $10^{-2}$, respectively. The theoretical optimal metrics, computed with $\operatorname{MMD}(\text{train}, \text{test})^2$, are used as the reference and represented by a light gray line.
Detailed results with higher precision are available in Appendix \ref{appendix:raw} for facilitating comparison.

\begin{table}[t]
    \centering
    \caption{Sample quality on large graphs. The mean ratios to the reference of the Degree, Cluster, Orbit, and Spectre MMD metrics are reported to enable a comprehensive comparison.}
    \resizebox{\columnwidth}{!}{
    \small
    \begin{tabular}{*{8}{l}}
    \toprule
     Class & Model    &  Degree$(10^{-3})$$\downarrow$ & Cluster $(10^{-2})$$\downarrow$ & Orbit $(10^{-2})$$\downarrow$ & Spectre $(10^{-3})$$\downarrow$ &  $\overline{\text{Ratio}}$ $\downarrow$ &  RBF $(10^{-2})$$\downarrow$   \\ \midrule 
     \rowcolor{lightgray}\cellcolor{white}\emph{Protein}& \emph{Reference} & 0.3          & 0.7 & 0.3 & 0.5 & \cellcolor{Gray}1.0& 1.4  \\
     Dense
    & GRAN   & 2.0 \hspace{.2cm} & 4.9 \hspace{.2cm} & 13 \hspace{.2cm}   & 5.1\hspace{.45cm} & \cellcolor{Gray}17 & --\hspace{0.55cm}  \\
    & DiGress      & $5.9 \spmb{0.1}$\hspace{-.3cm} & $10\spmb{1.4}$ \hspace{-.3cm} & $5.1\spmb{1.8}$  \hspace{-.3cm} & $2.9\spmb{0.5}$ \hspace{-.1cm} & \cellcolor{Gray}$14\spmb{2.3}$ &  $7.1\spmb{1.5}$\hspace{0.55cm}  \\
    Sparse & DRuM         & 1.9 \hspace{0.2cm} & 6.6 \hspace{0.2cm} & 3.5 \hspace{0.2cm} & 3.0\hspace{.45cm} & \cellcolor{Gray}8.4 &   --\hspace{0.55cm} \\
    &BiGG         & $\bm{1.0}$  \hspace{0.2cm} & $\bm{2.6}$  \hspace{0.2cm} & $2.3$  \hspace{0.2cm} &4.5 \hspace{0.45cm} &  \cellcolor{Gray}5.9 &   --\hspace{0.55cm} \\
    &HiGen        & 1.2 \hspace{0.2cm} & 4.4 \hspace{0.2cm} & 2.3\hspace{0.2cm} & 2.5\hspace{0.45cm} & \cellcolor{Gray}5.7 &   --\hspace{0.55cm} \\
    & GraphLE        & ${3.0}$ \hspace{0.2cm} & ${3.1}$ \hspace{0.2cm}  & ${\bm{0.5}}$ \hspace{0.2cm} & $\bm{1.3}$\hspace{.55cm} & \cellcolor{Gray}4.7 &  -- \hspace{0.4cm}  \\
    \rowcolor{Gray}
    \cellcolor{white}& SparseDiff   & $1.5\spmb{0.3}$\hspace{-.3cm} & $3.4\spmb{0.3}$\hspace{-.2cm} & $\bm{0.5}\spmb{0.8}$\hspace{-.3cm} & $\bm{1.4}\spmb{0.2}$\hspace{-.1cm} & \cellcolor{Gray}$\bm{3.6}\spmb{1.1}$ &   $\bm{3.8}\spmb{0.7}$ \\
    \midrule
    \rowcolor{lightgray}\cellcolor{white}\emph{Ego} & \emph{Reference} & 0.2             & 0.7             & 0.7  & 1.0 & \cellcolor{Gray}1.0 & 0.9  \\ 
    Dense & DiGress      & $8.9\spmb{1.6}$ \hspace{.2cm} & $5.4\spmb{0.4}$ \hspace{0.2cm} & $3.0\spmb{0.3}$ \hspace{0.2cm} & $19\spmb{3.2}$\hspace{.55cm} & \cellcolor{Gray}$19\spmb{3.1}$ & $3.4\spmb{0.8}$\hspace{.55cm}  \\
    Sparse &EDGE         & 58 \hspace{0.2cm} & 18 \hspace{0.2cm} & {5.2} \hspace{0.2cm} & --\hspace{.55cm} & \cellcolor{Gray}107 &  6.6\hspace{.55cm}  \\
    & HiGen        & ${47}$ \hspace{0.2cm} & ${\bm{0.3}}$ \hspace{0.2cm}  & ${3.9}$ \hspace{0.2cm} & --\hspace{.55cm} & \cellcolor{Gray}81 &  ${4.5}$ \hspace{0.4cm} \\
    \rowcolor{Gray}
    \cellcolor{white}
    &SparseDiff   & $\bm{3.7}\spmb{0.4}$ \hspace{-.3cm} & $3.2\spmb{0.1} $\hspace{-.15cm} & $\bm{2.0}\spmb{0.4}$ \hspace{-.15cm} & $\bm{5.6}\spmb{0.8}$ \hspace{-.1cm} & \cellcolor{Gray}$\bm{7.9}\spmb{0.9}$ & $\bm{2.6}\spmb{0.3}$\\
    \bottomrule
    \end{tabular}}
    \label{tab:proteinego}
\end{table}

\begin{table}
\centering
      \caption{
      Sample quality on synthetic graphs. The mean ratios to the reference of the Degree, Cluster and Orbit MMD metrics are reported to enable a comprehensive comparison.
      }
    \resizebox{\columnwidth}{!}{
    \begin{tabular}{ l *{5}{l} | *{5}{l}}
    \toprule
      Dataset    & \multicolumn{5}{c}{Stochastic block model} & 
       \multicolumn{5}{c}{Planar}\\ \midrule
     Model &  Degree$\downarrow$ & Cluster$\downarrow$ & Orbit$\downarrow$ & $\overline{\text{Ratio}}$ $\downarrow$ & V.U.N.$\uparrow$ &  Degree $\downarrow$ & Cluster $\downarrow$ & Orbit$\downarrow$ & $\overline{\text{Ratio}}$ $\downarrow$ & V.U.N.$\uparrow$ \\ \midrule
     \rowcolor{lightgray}
     \emph{Reference}     & 0.9\hspace{.55cm}  & 3.3\hspace{.4cm}  & 2.6\hspace{.4cm}     & \cellcolor{Gray}1.0    & 100\%\hspace{.25cm}&  0.2  \hspace{.2cm}  & 3.1 \hspace{.4cm} & 0.1    \hspace{.1cm}  & \cellcolor{Gray}1.0   & 100\%\\
     GRAN     & 5.5\hspace{.55cm}  & 5.8\hspace{.4cm}  & 7.9\hspace{.4cm} & \cellcolor{Gray}3.6 & 25\%\hspace{.25cm}&  \cellcolor{white}0.7  \hspace{.2cm}  & 4.3 \hspace{.4cm} & $\bm{0.1}$    \hspace{.1cm} & \cellcolor{Gray}2.0 & 0\%\\
     GG-GAN   & 3.5\hspace{.55cm}   & 7.0\hspace{.4cm}  & 5.9\hspace{.4cm} & \cellcolor{Gray}2.8 & 25\%\hspace{.25cm}&  \cellcolor{white}63\hspace{.3cm} & 118\hspace{.5cm} & 123\hspace{.2cm} & \cellcolor{Gray}528 & 0$\%$ \\ 
     SPECTRE  & 1.5\hspace{.55cm}   & 5.2\hspace{.4cm}  & $4.1$\hspace{.4cm} & \cellcolor{Gray}1.6 & 53\% &  \cellcolor{white}0.5  \hspace{.2cm}  & 7.9 \hspace{.4cm} & $\bm{0.1}$     \hspace{.1cm} & \cellcolor{Gray}2.0 &  25\%\\
     DiGress  & $1.7\spmb{0.1}$\hspace{.55cm} &$\bm{5.0}\spmb{0.1}$\hspace{.35cm} & $\bm{3.6}\spmb{0.4}$\hspace{.4cm} & \cellcolor{Gray}$\bm{1.6}\spmb{0.1}$ & $74\%\spmb{4}$
     & \cellcolor{white}$0.8 \spmb{0.0}$ \hspace{-.3cm} & $4.1\spmb{0.3}$& $0.5\spmb{0.0}$
     \hspace{-.3cm} & \cellcolor{Gray}$\bm{1.2}\spmb{0.4}$ & $76\%\spmb{1}$ \hspace{-.2cm} \\ 
     DruM   & $\bm{0.7}$\hspace{.55cm} & $\bm{4.9}$\hspace{.35cm} & $4.5$\hspace{.35cm} & \cellcolor{Gray}$3.6$ & $\bm{85}\%$ &  \cellcolor{white}0.5\hspace{.25cm} &$3.5$ \hspace{.4cm} &$\bm{0.1}$\hspace{.2cm} & \cellcolor{Gray}$1.5$ & $90\%$ \\
     HiGen  & 5.5\hspace{.55cm} & $5.8$\hspace{.35cm} & $7.9$\hspace{.35cm} & \cellcolor{Gray}3.6 & --\hspace{.3cm} & \cellcolor{white} --\hspace{.4cm} &--\hspace{.5cm} &--\hspace{.3cm} & \cellcolor{Gray}-- & --\hspace{.2cm} \\
     GraphLE  & 12\hspace{.45cm} & $5.2$\hspace{.35cm} & $6.7$\hspace{.35cm} & \cellcolor{Gray}5.8 & 45$\%$ &  \cellcolor{white}0.5\hspace{.25cm} &6.3\hspace{.45cm} &0.2\hspace{.15cm} & \cellcolor{Gray}2.2 & $\bm{95\%}$ \\
    \rowcolor{Gray}
     SparseDiff  & $\bm{1.6}\spmb{0.9}$ & $\bm{5.0}\spmb{0.1}$ \hspace{-.1cm} & $\bm{4.5}\spmb{0.9}$ \hspace{-.1cm} & \cellcolor{Gray}$\bm{1.7}\spmb{0.5}$ & $\bm{75}\%\spmb{10}$ \hspace{-.3cm}& 
     $\bm{0.3} \spmb{0.0}$ \hspace{-.3cm} & $\bm{3.2}\spmb{0.3}$& $\bm{0.1}\spmb{0.1}$
     \hspace{-.3cm} & \cellcolor{Gray}$\bm{1.2}\spmb{0.4}$ & $85\%\spmb{9}$ \hspace{-.2cm} \\ 
     \bottomrule
    \end{tabular}
    }
    \label{tab:sbm}
\end{table}

\paragraph{Results}
The results of large and synthetic graph generation are depicted in Tables \ref{tab:proteinego} and \ref{tab:sbm}. While dense models like DiGress \citep{vignac2022digress} perform excellently on SBM and planar graphs, those models struggle with larger datasets such as Ego and Protein due to quadratic space complexity. Dense graph diffusion models are limited to very small batch sizes (e.g., 2 on a 32GB GPU), resulting in slow training and poor convergence. In contrast, SparseDiff supports much larger batch sizes during training, thanks to the controllable sparsity parameter $\lambda$. Consequently, its performance matches previous dense and sparse models on mid-sized datasets like SBM and Planar and remains highly competitive with scalable models across various metrics on larger datasets such as Ego and Protein. Besides, we observe that our method consistently achieves the best results on the aggregated average ratio metrics, thereby validating its globally superior performance.

\subsection{Molecule Generation}
\label{subsec:molecule}

\begin{table}[t]
\centering
\caption{Molecule generation on QM9 with implicit hydrogens. Validity, uniqueness, and connectivity metrics are reported as percentages.}
\resizebox{\columnwidth}{!}{
\begin{tabular}{l *{4}{l} | l *{4}{l}}
    \toprule
    & \multicolumn{4}{c}{Dense Models} |& & \multicolumn{4}{c}{Sparse Models} \\
    \midrule
    Model & Valid $\uparrow$ & Unique $\uparrow$ & Conn. $\uparrow$ & FCD $\downarrow$ & Model & Valid $\uparrow$ & Unique $\uparrow$ & Conn. $\uparrow$ & FCD $\downarrow$ \\
    \midrule
    SPECTRE       & $87.3$ & $35.7$ & - & - &\cellcolor{white} GraphARM        & $90.3$ & - & - & $1.22$ \\
    GraphNVP      & $83.1$ & $99.2$ & - & - &\cellcolor{white} EDGE            & $99.1$ & $\bm{100}$ & - & $0.46$ \\
    GDSS          & $95.7$ & $98.5$ & - & $2.9$ &\cellcolor{white} HGGT           & $99.2$ & $95.7$ & - & $0.40$ \\
     \rowcolor{Gray}DiGress & $\bm{99.3}\spmb{0.0}$\hspace{-.05cm} & $95.9\spmb{0.2}$\hspace{-.05cm} & $99.4\spmb{0.2}\hspace{0.25cm}$ & $0.15 \spmb{0.01}$ 
     & SparseDiff & $\bm{99.2}\spmb{0.1} $\hspace{-.18cm}  & $96.4\spmb{0.1}$\hspace{-.18cm} & $\bm{99.8}\spmb{0.1}$ & $\bm{0.12}\spmb{0.00}$ 
     \\
    \bottomrule
\end{tabular}
}
\label{tab:qm9}
\end{table}

Given that our method behaves like dense models in the limit case where $\lambda=1$, it is expected to align with their performance on small graph datasets. We evaluate our approach using the QM9 and Moses molecular datasets, anticipating its performance comparable to that of dense models. The QM9 dataset \citep{wu2018moleculenet} features molecules with up to 9 heavy atoms, while the Moses benchmark \citep{polykovskiy2020molecular}, derived from ZINC Clean Leads, includes drug-sized molecules with extensive assessment tools. In QM9, we add formal charges as discrete node features during diffusion, similar to \citet{Vignac2023MiDiMG}, and apply the same to DiGress for consistency. We assess molecular performance by the proportion of connected graphs, validity of the largest connected component verified by RDKit, and uniqueness of over 10,000 molecules. Additionally, we use the Frechet ChemNet Distance (FCD) \citep{preuer2018frechet} to measure molecular similarity, excluding $0.96\%$ of invalid molecules for FCD analysis.

Table \ref{tab:qm9} demonstrates that SparseDiff consistently outperforms other scalable methods on the FCD metric, highlighting its effectiveness for small, structured graphs even without significant sparsity advantages. Additionally, SparseDiff achieves results comparable to the state-of-the-art dense model, DiGress, across other metrics.
Additional results for QM9 with explicit hydrogens and the Moses dataset (Tables \ref{tab:qm9withH} and \ref{tab:moses} in Appendices \ref{app:qm9h} and \ref{app:moses}) confirm that SparseDiff matches or exceeds the performance of the state-of-the-art model across various molecular datasets 
We further validate the robust performance of SparseDiff under different sparsity parameters $\lambda$ to confirm its stability (Tables \ref{tab:lambda} in Appendix \ref{app:add_exp_lambda}).

\subsection{Efficiency Analysis}
\label{subsec:efficiency}

\begin{minipage}[h]{0.47\textwidth}
\centering
\small
\vspace{-7pt} 
\vspace{0pt} 
\captionof{table}{Convergence comparison of graph diffusion models after 24 hours of training.}
\resizebox{\textwidth}{!}{
\begin{tabular}{l*{5}{l}}
\toprule
Model & Deg.$\downarrow$ & Clust.$\downarrow$ & Orbit$\downarrow$ &$\overline{\text{Ratio}}\downarrow$ & RBF$\downarrow$ \\
\midrule
EDP-GNN & $22$ & $36$ & $9.9$ & \cellcolor{Gray}59 & - \\
DiGress & $4.0$ & $4.9$ & $\bm{3.4}$ & \cellcolor{Gray}11 & $5.6$ \\
EDGE & $46$ & $18$ & $4.5$& \cellcolor{Gray}87 & $3.6$ \\
GraphLE & $58$ & $23$ & $4.2$ & \cellcolor{Gray}110 & - \\
SparseDiff & $\bm{2.3}$ & $\bm{4.7}$ & $3.6$ & \cellcolor{Gray}$\bm{7.8}$ & $\bm{3.5}$ \\
\bottomrule
\end{tabular}}
\label{tab:running_time_comparison}
\end{minipage}
\hfill
\begin{minipage}[h]{0.45\textwidth}
\centering
\captionof{figure}{Convergence comparison between DiGress and SparseDiff.}
\includegraphics[width=\textwidth]{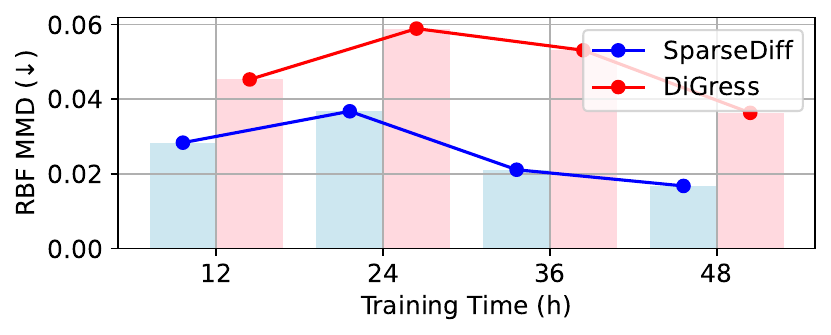}
\vspace{-15pt} 
\label{fig:convergence}
\end{minipage}

\paragraph{Training efficiency}

We compare SparseDiff's performance on the Ego dataset against other diffusion models, including dense models like EDP-GNN and DiGress, and sparse models such as EDGE and GraphLE. Table \ref{tab:running_time_comparison} shows that after 24 hours of training on a V100-32G machine, SparseDiff outperforms all metrics except for a minor increase in Orbit MMD compared to DiGress. The comparison of graph diffusion models' sampling speeds is further presented in Table \ref{tab:sampling_speed} in Appendix \ref{app:sampling_speed}. 
Figure \ref{fig:convergence} shows that SparseDiff has a significantly faster convergence speed compared to DiGress, achieving satisfactory results within two days. Notably, a SparseDiff model trained for 12 hours demonstrates an RBF MMD comparable to a DiGress model trained for 48 hours.

\begin{wraptable}{r}{0.45\textwidth} 
\small
\vspace{-18pt} 
\centering
\caption{Inference acceleration results. }
\resizebox{0.45\textwidth}{!}{
\begin{tabular}{ll*{5}{l}}
\toprule
Dataset & Steps & Deg.$\downarrow$ & Clust.$\downarrow$ & Orbit $\downarrow$&$\overline{\text{Ratio}}$ $\downarrow$& RBF$\downarrow$ \\
\midrule
\multirow{3}{*}{Ego} & $1000$ & $3.6$ & $3.1$ & $1.5$ &\cellcolor{Gray}8.2 & $2.0$ \\
                     & $500$  &$2.3$ & $2.9$ & $2.0$ & \cellcolor{Gray}6.2 & $2.1$  \\
                     & $200$  & $3.7$ & $3.1$ & $1.6$ &\cellcolor{Gray}8.4 & $2.3$ \\
\midrule
Dataset & Steps & Deg. & Clust. & Orbit&$\overline{\text{Ratio}}$ & V.U.N \\
\midrule
\multirow{3}{*}{Planar} & $1000$ & $0.3$ & $3.2$ & $0.1$&\cellcolor{Gray}1.2 & $85\%$ \\
                        & $500$ & $0.3$ & $3.4$  & $0.2$&\cellcolor{Gray}1.5 & $80\%$ \\
                        & $200$  & $0.5$ & $3.7$  & $0.4$&\cellcolor{Gray}2.6 & $69\%$ \\

\bottomrule
\end{tabular}}
\label{tab:acceleration_results}
\end{wraptable}

\paragraph{Inference efficiency}

We test the Ego and Planar datasets with different numbers of inference steps (1000, 500, and 200) after training with $T=1,000$ steps. The results, presented in Table \ref{tab:acceleration_results}, surprisingly demonstrate that even with a 5-fold increase in generation speed (down to 200 steps), our model keeps to be performing, and consistently outperforms most other dense and scalable models.

\section{Conclusion}
In this work, we introduce SparseDiff, a scalable discrete denoising diffusion model for graph generation. SparseDiff allows precise control over computational resources by predicting only a subset of edges at once. Experimental results demonstrate its superior and robust performance across graphs of all sizes, extending its applicability to tasks such as generating large molecules and community graphs
While SparseDiff meets most requirements of most scenarios, its scalability and ability to generate graphs with out-of-distribution node counts could be enhanced by incorporating a structured hierarchical approach. This enhancement promises to broaden its applicability in more complex graph generation tasks.


\bibliography{iclr2024_conference}
\bibliographystyle{iclr2024_conference}

\appendix

\newpage
\appendix
\onecolumn

\section{Proof of Lemma \ref{thm:tail}}
\label{app:math}

The lemma for a noisy graph with guaranteed sparsity comes directly from the proposition regarding the tail behavior of a binomial distribution \citep{desolneux2008estimating} as follows:

\begin{proposition} (Tail behavior of a binomial distribution) \label{thm:bino_tail} \\

Let $X_i, i=1,...l$ be independent Bernoulli random variables with parameter $0< p < \frac{1}{4}$ and let $S_l=\sum^l_{i=1}X_i$. Consider a constant $p< r< 1$ or a real function $p< r(l)< 1$. Then according to the Hoeffding inequality, $\mathcal{B}(l,k,p)=\mathbb{P}[S_l\geq k]$ satisfies:

\begin{equation}
    -\frac{1}{l} log \mathcal{P}[S_l\geq rl] \geq rlog\frac{r}{p} +(1-p) log\frac{1-r}{1-p}
\end{equation}
\end{proposition}

For sparse graphs, the edge ratio $r$ is clearly smaller than $\frac{1}{4}$. Consider then Bernoulli random variables with the parameter $r$ and a noised edge ratio $r<k<1$ with $l=n(n-1)/2$ (i.e. number of all node pairs in an undirected graph without self-loops) draws, and note the sampled `existing' edge number $S_{n(n-1)/2}$ as $m_t$, we have:


\begin{equation}
    log (\mathbb{P}[r_t=\frac{m_t}{n(n-1)/2}\geq k]) \leq -\frac{n(n-1)}{2} [k \operatorname{log} \frac{k}{r} + (1-r)\operatorname{log}\frac{1-k}{1-r}]
\end{equation}

\section{Model Architecture}
\label{app:arch}

Our sparse denoising network adopts the graph transformer architecture \citet{velivckovic2017graph}, featuring normalization, feed-forward, and attention layers. To handle sparse data, it incorporates the sparse attention mechanism \citep{shi2020masked} based on weighted message-passing layers and integrates enhancements from \citet{vignac2022digress}, such as PNA pooling layers \citep{corso2020principal} and FiLM layers \citep{perez2018film}.

Precisely, we introduce the FiLM layer and the PNA layer inside the model architecture to enhance its performance. Precisely, the FiLM layer is used to combine features at different scales, such as node and edge features. Specifically, given two features $\mM_1$ and $\mM_2$, and trainable parameters $\mW_1$ and $\mW_2$, the FiLM layer output is calculated as $\operatorname{FiLM}(\mM_1, \mM_2) = \mM_1 \mW_1 + (\mM_1 \mW_2) \odot \mM_2 + \mM_2$. As an illustration, within the convolutional layer, the graph feature $\mM_2$ is integrated with edge features $\mM_1$ to enhance predictions.
While PNA layer is used as a specialized pooling layer to obtain information from different dimensions of a specific feature. Given the feature X and trainable parameter $\mW$, $\operatorname{PNA}(\mX) = \operatorname{cat}(\operatorname{max}(\mX), \operatorname{min}(\mX),  \operatorname{mean}(\mX), \operatorname{std}(\mX)) ~ \mW $. For example, node features $\mX$ are forwarded to a PNA layer for extracting global information across different scales, which is subsequently added to the graph feature to enhance its representation.

Finally, we enrich the message-passing graph with additional encodings, such as the graph Laplacian and cycle counts, to enhance structural and positional information (detailed in Appendix \ref{app:encoding}). These encodings can only be computed effectively when the noisy graphs are sparse, which is another significant advantage of discrete diffusion models using marginal transitions. It is worth noting that not all these encodings can be computed in sub-quadratic time. However, in practice, this does not pose an issue as they are not used for back-propagation, which arises as the primary complexity bottleneck during training. For instance, for the large Protein dataset, computing these encodings is five times faster than the forward pass itself. Nevertheless, on very larger graphs, expensive ones should not be computed.

\subsection{Additional Encodings}
\label{app:encoding}

During training,  we augment model expressiveness with additional encodings. To make things clear, we divide them into encodings for edges, nodes, and for graphs.

\paragraph{Encoding for graphs}
We first incorporate graph eigenvalues, known for their critical structural insights, and cycle counts, addressing message-passing neural networks' inability to detect cycles \citep{chen2020can}. The first requires $n^3$ operations for matrix decomposition, the second requires $n^2$ for matrix multiplication, but both are optional in our model and do not significantly limit scalability even with graphs up to size $500$.
In addition to the previously mentioned structural encodings, we integrate the degree distribution to enhance the positional information within the graph input, which is particularly advantageous for graphs with central nodes or multiple communities. Furthermore, for graphs featuring attributed nodes and edges, the inclusion of node type and edge type distributions also provides valuable benefits.

\paragraph{Encoding for nodes}
At the node level, we use graph eigenvectors, which are fundamental in graph theory, offering valuable insights into centrality, connectivity, and diverse graph properties.

\paragraph{Encoding for edges}
To aid in edge label prediction, we introduce auxiliary structural encodings related to edges. These include the shortest distance between nodes and the Adamic-Adar index. The former enhances node interactions, while the latter focuses on local neighborhood information. Due to computational constraints, we consider information within a 10-hop radius, categorizing it as local positional information.

\paragraph{Molecular information}
In molecular datasets, we augment node features by incorporating edge valency and atom weights. Additionally, formal charge information is included as an additional node label for diffusion and denoising during training, as formal charges have been experimentally validated as valuable information \citep{Vignac2023MiDiMG}.

\begin{figure}[t]
     \centering
     \includegraphics[width=0.99\linewidth]{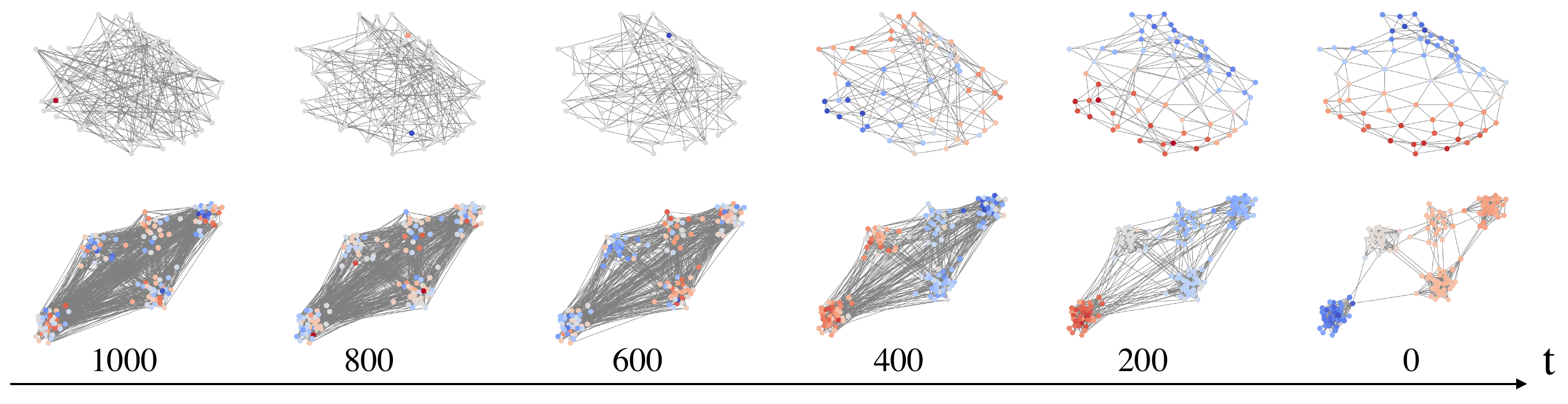}
     \caption{Visualization of iterative generation for Planar and SBM graphs.}
     \label{fig:iter_sampling}
\end{figure}

\section{Experimental setup}
\label{app:setup}

In our experimental setup, we utilize a single V100-32G GPU machine, which is particularly prone to scalability issues, to demonstrate that our method allows users with limited GPU resources to effectively train on larger graphs. Detailed specifications regarding workers, memory allocation, execution time, and optimizers are meticulously indicated in the configuration details provided in our code.

As for dataset splits, we adhere to the framework established by DiGress. Specifically, for the QM9 dataset, we implement a split comprising 100k molecules for training, 20k for validation, and 13k for evaluating likelihood on the test set. For the Planar, SBM, and Protein datasets, employing a seed of 1234, we randomly assign $20\%$ of the graphs to testing, while $80\%$ of the remaining graphs are utilized for training, and $20\%$ for validation. For the Ego dataset, to ensure consistency with previous methods and a fair comparison, we maintain a split of $80\%$ for training and $20\%$ for testing, with $20\%$ of the training set additionally used for validation purposes. All configuration details are comprehensively documented in the code provided.

\subsection{MMD metrics}
In our research, we carefully select specific metrics tailored to each dataset, with a primary focus on four widely recognized Maximum Mean Discrepancy (MMD) metrics. These metrics use the total variation (TV) distance, as detailed in \citet{martinkus2022spectre}. They encompass node degree (Deg), clustering coefficient (Clus), orbit count (Orb), and graph spectra (Spec). The first three local metrics compare the degree distributions, clustering coefficient distributions, and the occurrence of all 4-node orbits within graphs between the generated and training samples. Additionally, the comparison of graph spectra is realized by computing the eigenvalues of the normalized graph Laplacian, providing complementary insights into the global properties of the graphs.

\subsection{Statistics of the datasets}
\label{app:stats}
\begin{table}[H]
\centering
\caption{Statistics for the datasets employed in our experiments.}
\begin{tabular}{l r r r r r } \toprule
Name & Graph number & Node range & Edge range & Edge Ratio ($\%$)  & $\lambda$ ($\%$)\\ \midrule
QM9         & 133,885 & [2,9]       & [2, 28]        &  [20, 56]    & 100 \\
QM9(H)      & 133,885 & [3, 29]     & [4, 56]        &  [7.7, 44]   & 50 \\
Moses       & 1,936,962 & [8, 27]   & [14, 62]       &  [8.0, 22]   & 50 \\ 
Planar      & 200     & [64, 64]    & [346, 362]     &  [8.4, 8.8]  & 50 \\ \midrule 
SBM         & 200     & [44, 187]   & [258, 2258]    &  [6.0, 17]   & 25 \\
Ego         & 757     & [50, 399]   & [112, 2124]    &  [1.2, 11]   & 10 \\
Protein     & 918     & [100, 500]  & [372, 3150]    &  [8.9, 6.7]  & 10 \\
\bottomrule
\end{tabular}\vspace{0.1cm}
\label{tab:stats}
\end{table}

To provide a more comprehensive overview of the various scales found in `existing' graph datasets, we present here key statistics for them. These statistics encompass the number of graphs, the range of node numbers, the range of edge numbers, the edge ratio for `existing' edges, and the sparsity parameter $\lambda$ used for training, i.e. the proportion of `existing' edges among all node pairs. In our consideration, we focus on undirected graphs. Therefore, when counting edges between nodes $i$ and $j$, we include the edge in both directions.

\subsection{Raw results}\label{appendix:raw}

\begin{table}[H]
    \centering
    \caption{Raw results on the SBM, Planar, Protein, and Ego datasets.}
    \label{tab:large_num}
    \resizebox{\columnwidth}{!}{
    \begin{tabular}{l l l l l l l l}
    \toprule
     Model    &  Deg. (e-3)$\downarrow$ & Clust. (e-2)$\downarrow$ & Orbit (e-2)$\downarrow$ & Spec. (e-3)$\downarrow$ & FID$\downarrow$  & \hspace{-.6cm}RBF MMD (e-2)$\downarrow$ \hspace{.6cm}\\ \midrule 
    \multicolumn{2}{l}{\emph{SBM}} \\
    Training set & 0.85            & 3.32             & 2.55 & 2.74 & 1.37 & 3.23\hspace{.6cm} \\
    SparseDiff   & $1.57\spmb{0.91}$ & $5.04\spmb{0.06}$ & $4.51\spmb{0.90}$ & $6.68\spmb{2.04}$ & $4.55\spmb{2.01}$ & $4.98\spmb{0.06}$ \\
    \multicolumn{2}{l}{\emph{Planar}} \\
    Training set & 0.19           & 3.10              & 0.05 & 3.82 & 1.57 & 8.89 \\
    SparseDiff & $0.32\spmb{0.01}$ & $3.25\spmb{0.35}$ & $0.09\spmb{0.08}$ & $6.99\spmb{0.92}$ & $2.94\spmb{3.15}$ & $9.84\spmb{0.91}$ \\
    \multicolumn{2}{l}{\emph{Protein}} \\
    Training set & 0.32          & 0.68             & 0.32 & 0.49 & 1.36 & 1.37\hspace{.6cm} \\
    SparseDiff & $1.45\spmb{0.30}$ & $3.35\spmb{0.33}$ & $0.53\spmb{0.78}$ & $1.35\spmb{0.16}$ & $5.97\spmb{1.07}$ & $3.77\spmb{0.65}$ \\
    \multicolumn{2}{l}{\emph{Ego}} \\
    Training set & 0.16             & 0.71             & 0.69  & 0.98 & 0.07 & 0.86\hspace{.6cm} \\
    SparseDiff & $3.70\spmb{0.44}$  & $3.18\spmb{0.10}$ & $1.98\spmb{0.42}$ & $5.63\spmb{0.80}$ & $4.84\spmb{1.56}$ & $2.60\spmb{0.31}$ \\
    \bottomrule
    \end{tabular}
    }
\end{table}

To ease comparison with other methods, Table \ref{tab:large_num} provides the raw numbers (not presented as ratios) for the SBM, Planar, Ego, and Protein datasets. Note that this table contains the FID metrics from \citet{thompson2022evaluation}, which we did not include in the main text. The reason is that we found this metric to be very brittle, with some evaluations giving a very large value that would dominate the mean results. Besides, we have identified a discrepancy in the Spectre metrics reported in the study by \citet{martinkus2022spectre} when computed under non-parallel computation. We thus provide the updated values for reference and use the updated value for ratio calculation in Table \ref{tab:sbm} and in Table \ref{tab:proteinego}. 
The checkpoints needed to reproduce the results mentioned can be accessed via the provided URL. The code is available at \url{https://github.com/qym7/SparseDiff}. The checkpoints for reproducing the results above can be found at \url{https://drive.switch.ch/index.php/s/1hHNVCb0ylbYPoQ}.

\section{Additional experiments}
\label{app:add_exp}

\subsection{Training with graph of over 1000 nodes}
\label{app:single_graph}

\begin{table}[H]
    \centering
    \caption{Large graph generation on the Facebook dataset. Triangles and squares are abbreviated as `tri' and `squ' in the table, while PLE represents power law exp.}
    \label{tab:large}
    \resizebox{\columnwidth}{!}{
    \begin{tabular}{l l l l l l l l l l l}
    \toprule
    Model&num nodes&num edges&num tri&num squ&max deg&clust&assort&PLE&CPL\\
    \midrule
    Real&1045&27,755&446,846&34,098,662&1044&0.57579&$-$0.02543&1.28698&1.94911 \\
    \midrule
    SaGess&1043&\textbf{27,758}&429,428&35,261,545&999&\textbf{0.52098}&$-$0.01607&1.29003&2.00800 \\
    SparseDiff&\textbf{1045}&27,763&\textbf{446,819}&\textbf{34,095,513}&\textbf{1044}&0.43310&$\bm{-.02536}$&\textbf{1.28687}&\textbf{1.94921} \\
     \bottomrule
    \end{tabular}
    }
    \label{tab:training_time}
\end{table}

We trained on the largest graph with 1045 nodes from the Facebook dataset, following the SaGess \citep{limnios2023sagess} setting. SparseDiff was evaluated using SaGess metrics as a reference. However, employing the same graph for both training and evaluating test metrics entails potential risks, as perfect metrics may result from overfitting to the training graph. In the provided table, we present SaGess-RW, demonstrating the best results among the three proposed SaGess models. Notably, SaGess generates small graphs and concatenates them to meet the required number of edges, while SparseDiff generates a single large graph based on the specified node count. This explains SparseDiff's advantage in the `num nodes' metric and SaGess's advantage in the `num edges' metric. Furthermore, SparseDiff closely aligns with real data statistics, except for the clustering coefficient, showcasing not only its scalability up to 1000 nodes but also its strong performance on such single-graph datasets.

\subsection{QM9 with explicit hydrogens}
\label{app:qm9h}

\begin{table}[H]
    \centering
    \caption{Unconditional generation on QM9 with explicit hydrogens. On small graphs such as QM9, sparse models are not beneficial, but SparseDiff still achieves very good performance.}     \label{tab:qm9withH}
    \begin{tabular}{l c c c c c c c c c} \toprule
Model     & Connected &  Valid$\uparrow$\hspace{0.3cm} & Unique$\uparrow$\hspace{0.2cm} & Atom stable$\uparrow$ & Mol stable$\uparrow$ \\ \midrule
DiGress  & --  & $95.4$ & $\bm{97.6}$ & $98.1$ & $79.8$ \\
DiGress + charges & \textbf{98.6} & 97.7 & $96.9$ \hspace{-.1cm} & 99.8 & 97.0 \\
    SparseDiff(ours)& $98.3\spmb{.08}$\hspace{-.6cm} &$\bm{97.9}\spmb{.13}$ \hspace{-.7cm} & $97.4\spmb{.10}$ \hspace{-.7cm} & -  & -\\
\bottomrule
    \end{tabular}
\end{table}

We additionally report the results for QM9 with explicit hydrogens in Table \ref{tab:qm9withH}. Having explicit hydrogens makes the problem more complex because the resulting graphs are larger. We observe that SparseDiff achieves better validity than DiGress and has comparable results on other metrics when both are utilizing charges.

\subsection{Moses benchmark evaluation}
\label{app:moses}

\begin{table}[H]
    \centering
    \caption{Mean and standard deviation across 5 samplings on the Moses benchmark. SparseDiff has a similar performance to DiGress, despite a shorter training time.}
    \label{tab:moses}
    \resizebox{\columnwidth}{!}{
    \begin{tabular}{l r r r r r} \toprule
    Model & Connected $\uparrow$ &  Valid $(\%) \uparrow$ & Unique $(\%) \uparrow$ & Novel $(\%) \uparrow$ & Filters $(\%)\uparrow$  \\ 
    \midrule
    GraphINVENT & --\hspace{.55cm} & $\bm{96.4} \hspace{.43cm}$& $99.8\hspace{.43cm}$ & --\hspace{.55cm} & $95.0\hspace{.43cm}$ \\
    DiGress & --\hspace{.55cm} & $85.7\hspace{.4cm}$ & $\bm{100.0}\hspace{.43cm}$ & $95.0\hspace{.43cm}$ & $\bm{97.1}\hspace{.43cm}$ \\
    SparseDiff  & $94.8 \spmb{.1}$ & $84.7 \spmb{.2}$ & $\bm{100.0} \spmb{.0}$ & $\bm{95.1} \spmb{.1}$ & $\bm{97.0} \spmb{.2}$  \\
    \midrule
     Model & FCD $\downarrow$\hspace{.2cm} & SNN $ (\%) \uparrow$ & Scaf $(\%) \uparrow$ & Frag $(\%) \uparrow$ & IntDiv $(\%) \uparrow$ \\ \midrule
     GraphINVENT & $1.22\hspace{.55cm}$ & $\bm{53.9}\hspace{.4cm}$ & $12.7\hspace{.55cm}$ & $98.6\hspace{.43cm}$ & $\bm{85.7}\hspace{.43cm}$\\
     DiGress & $\bm{1.19}\hspace{.55cm}$ & $52.2\hspace{.4cm}$ & $\bm{14.8}\hspace{.55cm}$ & $99.6\hspace{.43cm}$ & $85.3\hspace{.43cm}$\\
     SparseDiff & $1.28 \spmb{.01}$ & $52.2\spmb{.0}$ & $\bm{15.5} \spmb{1.3}$ & $\bm{99.8} \spmb{.0}$ & $85.4 \spmb{.0}$\\ \midrule
     Model &  IntDiv2  $ (\%) \uparrow$ & logP $(e^{-2}) \downarrow$ & SA$(e^{-2}) \downarrow$\hspace{.2cm} & QED $(e^{-3}) \downarrow$ & Weight $(\%)\downarrow$ \\ \midrule
     GraphINVENT & $\bm{85.1}\hspace{.43cm}$ & $\bm{0.67}\hspace{.3cm}$ & $4.5\hspace{.43cm}$ & $\bm{0.25}\hspace{.55cm}$ & $16.1\hspace{.43cm}$\\
     DiGress & $-\hspace{.43cm}$ & $3.4\hspace{.43cm}$ & $\bm{3.6}\hspace{.43cm}$ & $2.91\hspace{.55cm}$ & $\bm{1.42}\hspace{.4cm}$\\
     SparseDiff & $84.8 \spmb{.0}$ & $3.0 \spmb{.3}$ & $5.4 \spmb{.2}$ & $1.21 \spmb{.21}$ & $5.58 \spmb{.15}$\\
    \bottomrule
    \end{tabular}
    }
\end{table}

Moses is an extensive molecular dataset with larger molecular graphs than QM9, offering a much more comprehensive set of metrics. While autoregressive models such as GraphINVENT are recognized for achieving higher validity on this dataset, both SparseDiff and DiGress exhibit advantages across most other metrics. Notably, SparseDiff closely aligns with the results achieved by DiGress, affirming the robustness of our method on complex datasets.

\subsection{Sampling Speed Comparison}\label{app:sampling_speed}

\begin{table}[H]
\centering
\caption{Sampling speed for generating 8 Ego graphs.}
\begin{tabular}{ccccccc}
\toprule
Model & EDP-GNN & DiGress & EDGE & GraphLE & SparseDiff & SparseDiff (200 steps) \\
\midrule
Time (min) & $5$ & $32$ & $2$ & $20$ & $28$ & $5$ \\
\bottomrule
\end{tabular}
\label{tab:sampling_speed}
\end{table}
The speed of generating 8 Ego graphs is demonstrated in Table \ref{tab:sampling_speed}. Notably, for GraphLE, the batch size is constrained to $2$ due to its substantial memory requirements. An additional column labeled ``SparseDiff (200 steps)'' represents the sampling time after reducing the inference steps from 1,000 to 200 through acceleration strategies. The table illustrates that SparseDiff maintains comparable speed to dense models without significant compromise on space efficiency and can be significantly accelerated during sampling.

\subsection{Influence of the sparsity parameter}
\label{app:add_exp_lambda}

\begin{table}[H]
    \centering
    \caption{Unconditional generation on QM9 under different sparsity parameters $\lambda$.}
    \label{tab:lambda}
    \begin{tabular}{c c c c c c c c c} 
    \toprule
    $\lambda$ & Valid$\uparrow$ & Unique$\uparrow$ & Connected$\uparrow$  & FCD$\downarrow$ \\ 
    \midrule
    $100\%$ & $99.23\spmb{0.06}$ & $96.37\spmb{0.13}$ & $99.76\spmb{0.06}$ & $0.117\spmb{0.004}$ \\
    $50\%$ & $99.12\spmb{0.05}$ & $96.80\spmb{0.18}$ & $99.61\spmb{0.05}$ & $0.107\spmb{0.007}$ \\
    $25\%$ & $99.16\spmb{0.06}$ & $96.54\spmb{0.19}$ & $99.59\spmb{0.06}$ & $0.119\spmb{0.006}$ \\
    $10\%$ & $99.11\spmb{0.09}$ & $96.89\spmb{0.15}$ & $99.61\spmb{0.01}$ & $0.105\spmb{0.004}$ \\
    \bottomrule
    \end{tabular}
\end{table}

Table \ref{tab:lambda} above shows the results of unconditional generation on the QM9 dataset under different sparsity parameters ($\lambda$). The performance of SparseDiff remains consistent across various $\lambda$ values, with metrics for connectivity, validity, uniqueness, and Frechet ChemNet Distance (FCD) showing minimal changes from $100\%$ to $10\%$ $\lambda$. For example, the validity metric stays between $99.12\%$ and $99.23\%$, and uniqueness ranges from $96.37\%$ to $96.89\%$. We remark that, for $\lambda$ values of $10\%$ and $25\%$, the model was trained for twice as many epochs since the cross-entropy loss considers fewer edges per epoch, necessitating more epochs for convergence. Despite this, all models exhibit consistent performance across different $\lambda$ values after convergence, which highlights the robustness and stability of SparseDiff in generating high-quality molecular graphs.

\subsection{Ablations}
\label{app:ablations}

This part presents 2 ablation experiments that motivate our approach. 
SparseDiff builds upon an experimental observation and a hypothesis.
Firstly, our experiments demonstrate that relying solely on node features for link prediction yields unsatisfactory results. This observation encouraged us to design the message-passing graph that contains all edges to be predicted (i.e. query edges) as the message-passing graph to directly obtain their edge features.
Secondly, we hypothesized that preserving the same distribution of edge types as observed in dense graphs for loss calculation is advantageous for training. 
This hypothesis necessitates the sampling of query edges within each graph in a batch of graphs with varying sizes, thereby introducing increased complexity to the algorithm design process.

\subsubsection{Link Prediction}

\begin{table}[H]
    \centering
    \caption{Influence of including edges features for edge prediction.}
    \label{tab:link}
    \resizebox{\columnwidth}{!}{
    \begin{tabular}{l |  l l l l l l l}
    \toprule
     Model    &  Deg $\downarrow$ & Clus $\downarrow$ & Orb$\downarrow$ & Spec$\downarrow$ & FID$\downarrow$ & RBF MMD$\downarrow$ \\ \midrule 
     Link Pred & 0.0043 & 0.0721 & \textbf{0.0275} & 0.0344 & 1.51e6 & \textbf{0.0315} \hspace{.3cm} \\
     SparseDiff  & $\textbf{0.0019}\spmb{.00}$  & $\bm{0.0537}\spmb{.00}$ & $0.0299\spmb{.00}$ & $\bm{0.0050}\spmb{.00}$ & $\bm{16.1}\spmb{12.9}$ & $0.0483\spmb{.01}$ \\
     \bottomrule
    \end{tabular}
    }
\end{table}

In this experiment, we intentionally avoided using easily learnable molecular datasets that come with rich supplementary encodings. Instead, we chose to conduct the experiments on a large dataset, namely Ego, to assess their performance. 
In Table \ref{tab:link}, a model that does not specifically include edge features for edge prediction performs much worse on all metrics except on RBF MMD and orbit. 
This observation shows that, despite that a model can also leverage the information of `existing' edges into node features, the lack of edge features during training deteriorates significantly models' performance. 

\subsubsection{Query edges with proper distribution}

\begin{table}[h]
    \centering
    \caption{Influence of edge loss distribution on EGO dataset.}
    \label{tab:ego_abl3}
    \resizebox{\columnwidth}{!}{
    \begin{tabular}{l |  l l l l l l l}
    \toprule
     Loss based on    &  Deg $\downarrow$ & Clus $\downarrow$ & Orb$\downarrow$ & Spec$\downarrow$ & FID$\downarrow$ & RBF MMD$\downarrow$ \\ \midrule 
     Comp graph & 0.0021 &0.0566 &$\bm{0.0270}$ & 0.0100 & 28.2 & $\bm{0.0396}$  \hspace{.3cm} \\
     Query graph  & $\textbf{0.0019}\spmb{.00}$  & $\bm{0.0537}\spmb{.00}$ & $0.0299\spmb{.00}$ & $\bm{0.0050}\spmb{.00}$ & $\bm{16.1}\spmb{12.9}$ & $0.0483\spmb{.01}$ \\
     \bottomrule
    \end{tabular}
    }
\end{table}


In order to emphasize the importance of preserving the edge distribution when computing losses, we conduct an experiment where we assess the performance of a model trained using all message-passing edges as opposed to solely using query edges. The former results in an increased emphasis on `existing' edges during training compared to SparseDiff.
Similarly, we use the Ego dataset for initial experiments. Table \ref{tab:ego_abl3} shows that using edges of the message-passing graph $G_m$ results in worse performance on most of the metrics, which indicates the importance of keeping a balanced edge distribution for loss calculation.

\section{Social Impact}
\label{app:social}

This paper aims to advance the field of graph generation.
Our method enhances the training of larger graphs on limited GPU resources, enabling advancements in fields such as computational biology by allowing biologists to work with expansive protein graphs. It is also valuable in social network analysis and financial fraud detection by efficiently managing large-scale graphs, aiding in the understanding of complex patterns. While our work may have various societal implications as mentioned above, none are particularly harmful and thus do not require specific mention at this time.

\section{Visualization}
\label{app:visual}

\begin{figure*}[H]
    \centering
    \begin{subfigure}[b]{0.9\linewidth}
         \centering
         \includegraphics[width=0.9\linewidth]{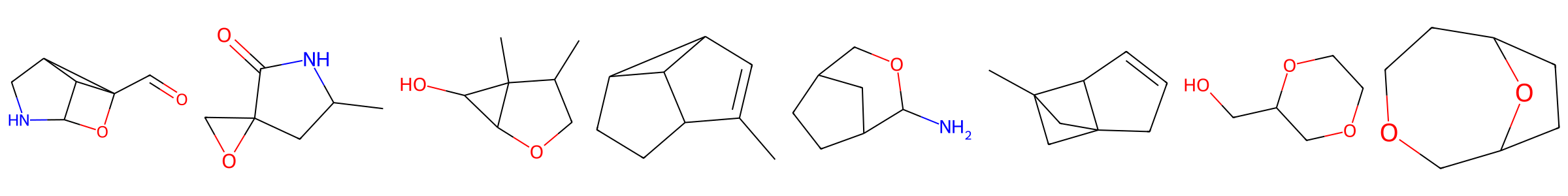}
         \caption{Training graphs.}
    \end{subfigure}
    \hfill
    \begin{subfigure}[b]{0.9\linewidth}
         \centering
         \includegraphics[width=0.9\linewidth]{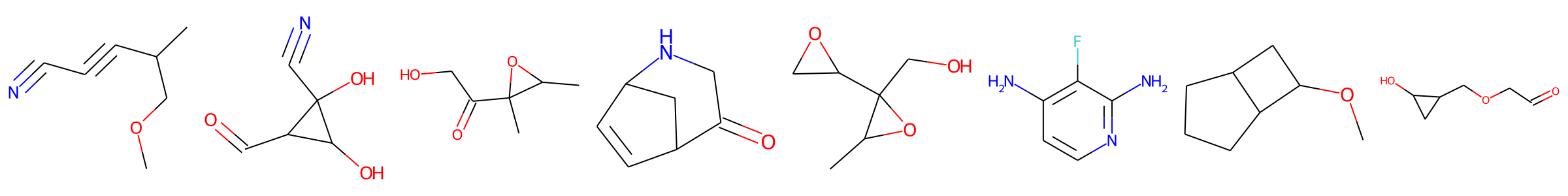}
         \caption{Generated graphs.}
    \end{subfigure}
    \hfill
    \caption{Visualization for QM9 dataset with implicit hydrogens.
    }
    \label{fig:vis_qm9}
\end{figure*}

\begin{figure*}[H]
    \centering
    \begin{subfigure}[b]{0.9\linewidth}
         \centering
         \includegraphics[width=0.9\linewidth]{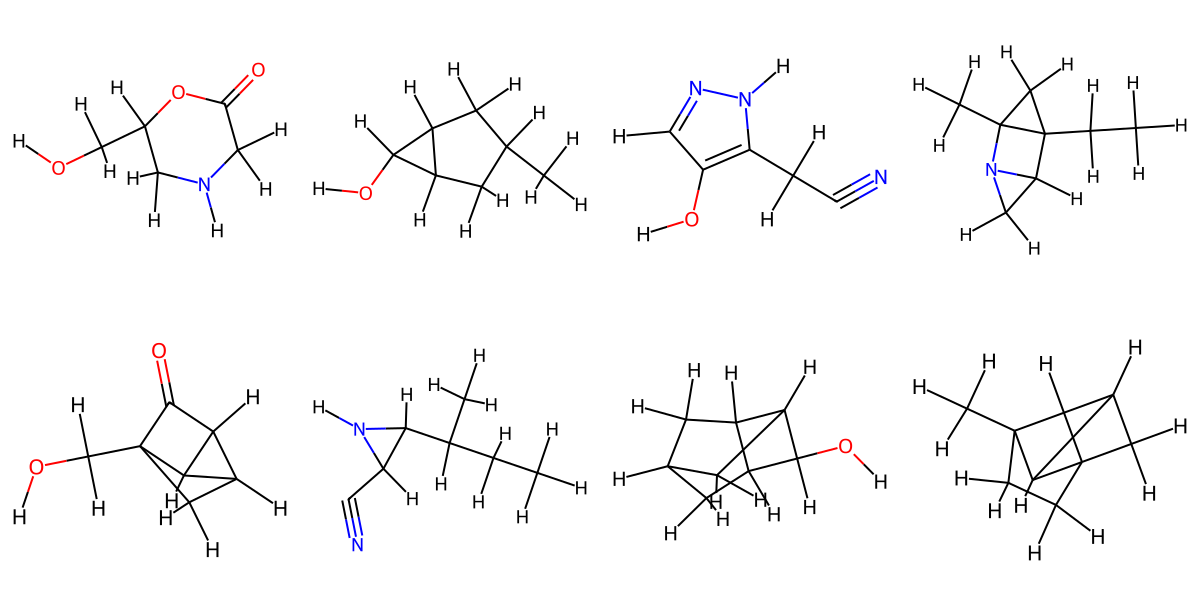}
         \caption{Training graphs.}
    \end{subfigure}
    \hfill
    \begin{subfigure}[b]{0.9\linewidth}
         \centering
         \includegraphics[width=0.9\linewidth]{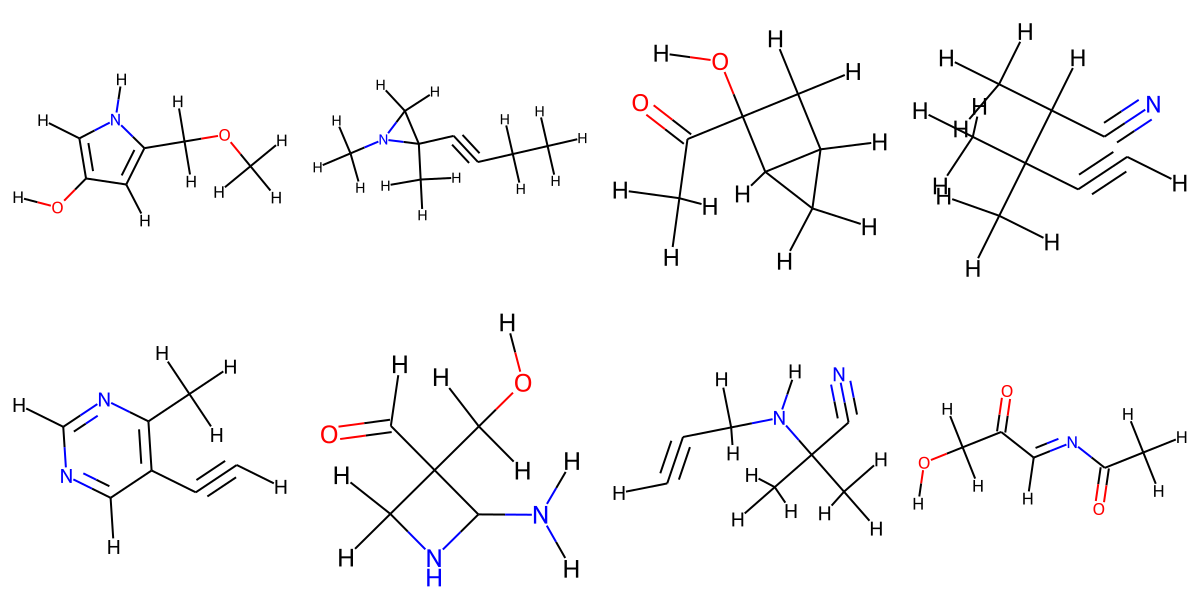}
         \caption{Generated graphs.}
    \end{subfigure}
    \hfill
    \caption{Visualization for QM9 dataset with explicit hydrogens.
    }
    \label{fig:vis_qm9h}
\end{figure*}

\begin{figure}[H]
    \centering
    \begin{subfigure}[b]{0.99\linewidth}
         \centering
         \includegraphics[width=0.9\linewidth]{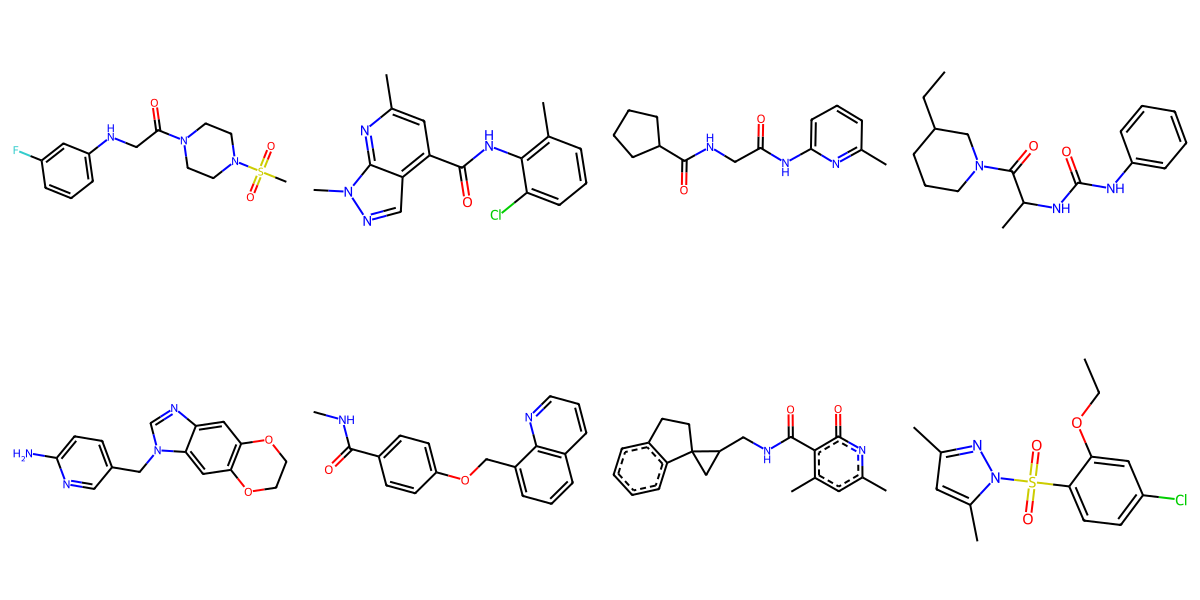}
         \caption{Training graphs.}
    \end{subfigure}
    \hfill
    \begin{subfigure}[b]{0.99\linewidth}
         \centering
         \includegraphics[width=0.9\linewidth]{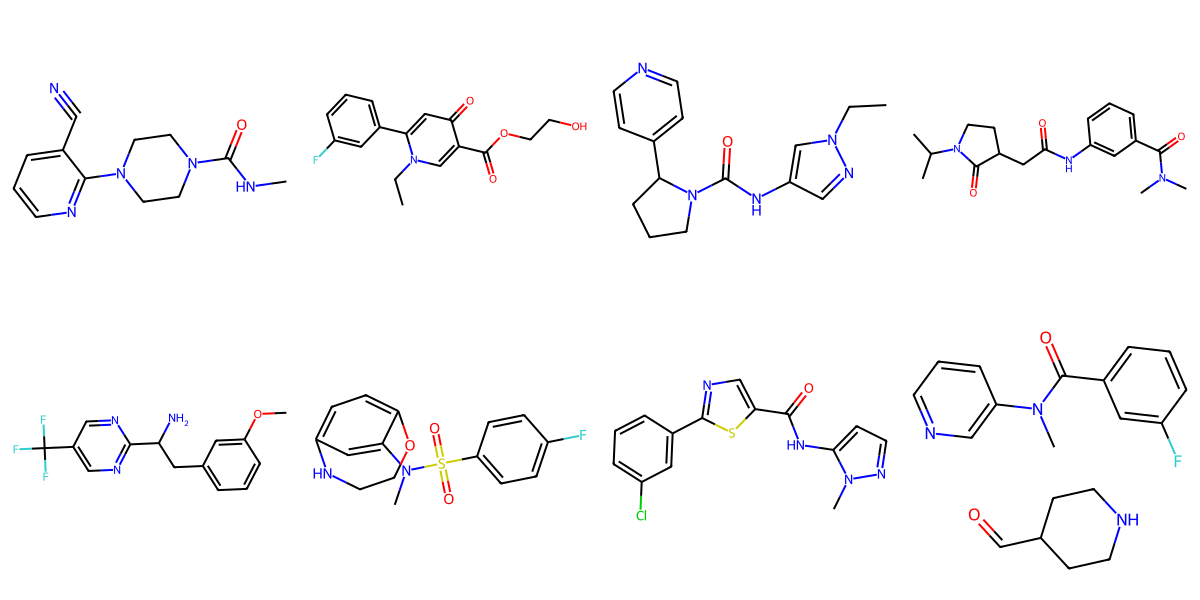}
         \caption{Generated graphs.}
    \end{subfigure}
    \hfill
    \caption{Visualization for Moses dataset.
    }
    \label{fig:vis_moses}
\end{figure}

\begin{figure}[H]
    \centering
    \begin{subfigure}[b]{0.99\linewidth}
         \centering
         \includegraphics[width=0.9\linewidth]{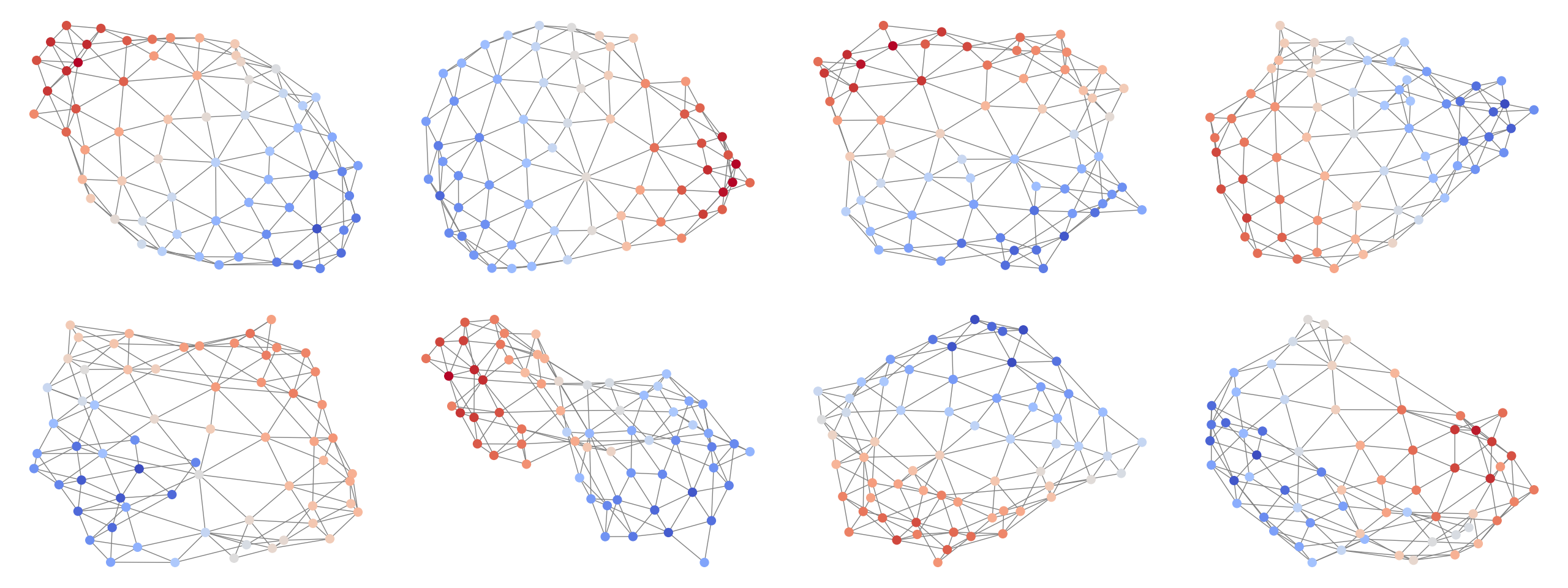}
         \caption{Training graphs.}
    \end{subfigure}
    \hfill
    \begin{subfigure}[b]{0.99\linewidth}
         \centering
         \includegraphics[width=0.9\linewidth]{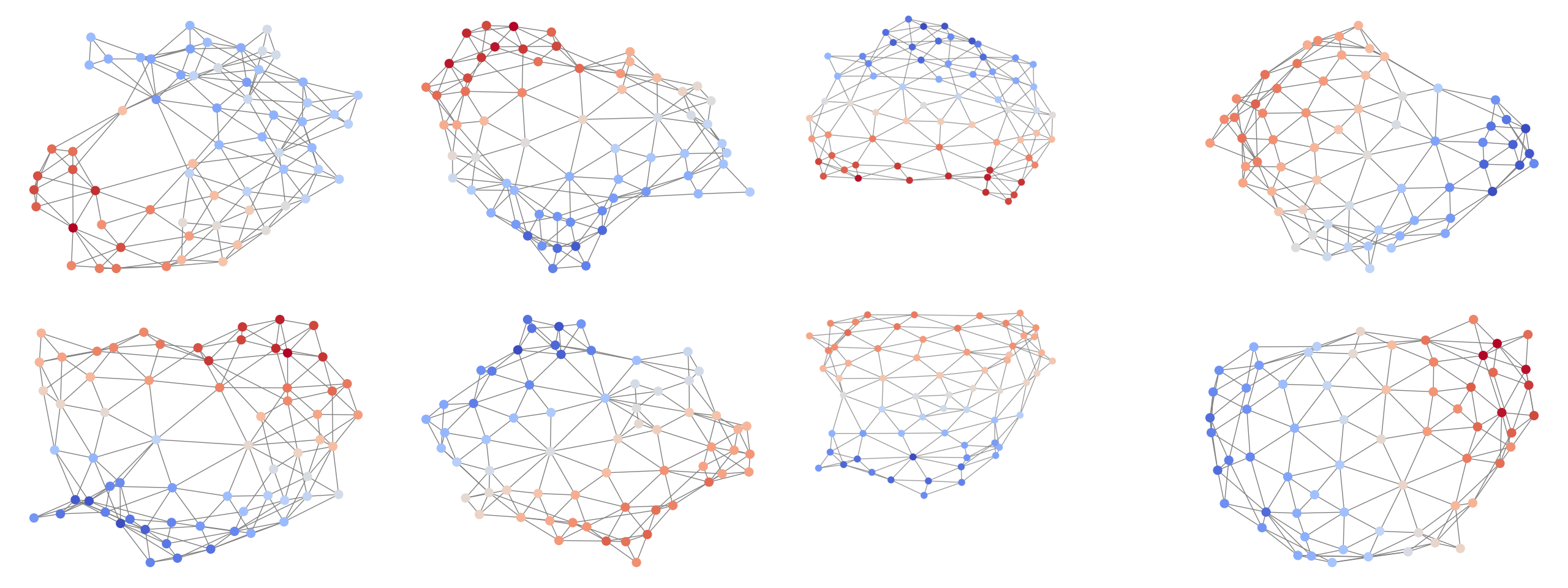}
         \caption{Generated graphs.}
    \end{subfigure}
    \hfill
    \caption{Visualization for Planar dataset.
    }
    \label{fig:vis_planar}
\end{figure}

\begin{figure}[H]
    \centering
    \begin{subfigure}[b]{0.99\linewidth}
         \centering
         \includegraphics[width=0.9\linewidth]{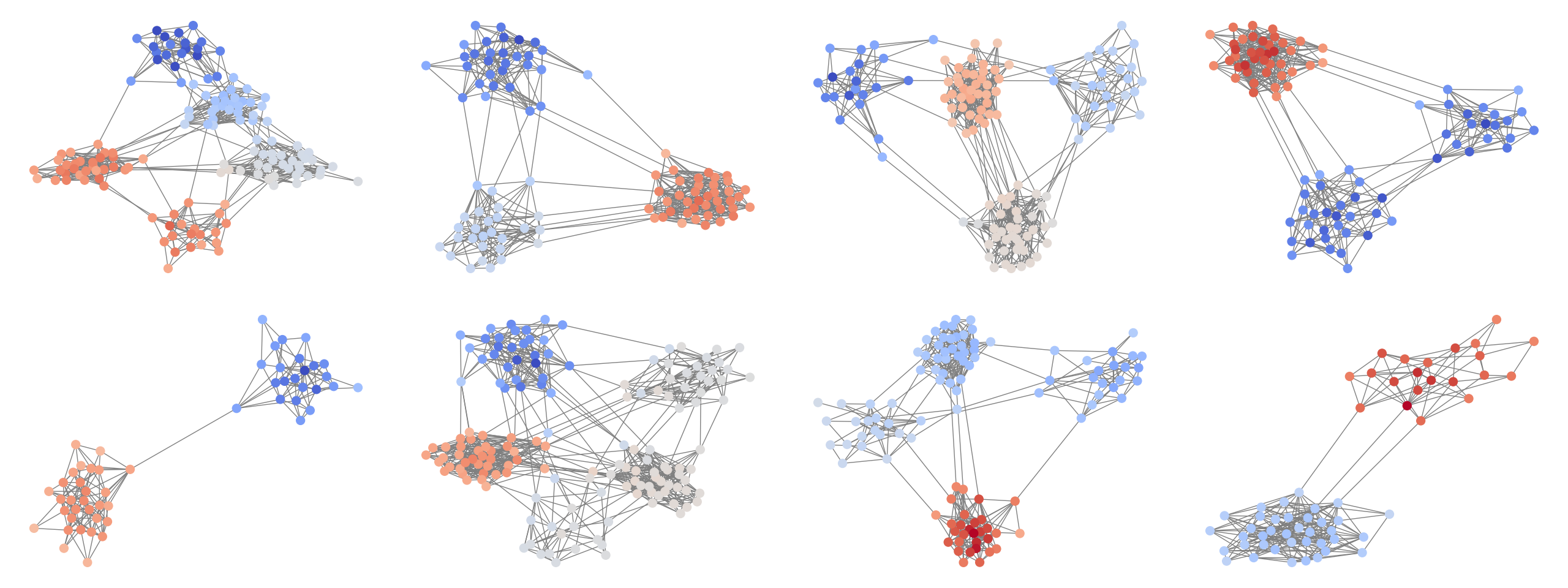}
         \caption{Training graphs.}
    \end{subfigure}
    \hfill
    \begin{subfigure}[b]{0.99\linewidth}
         \centering
         \includegraphics[width=0.9\linewidth]{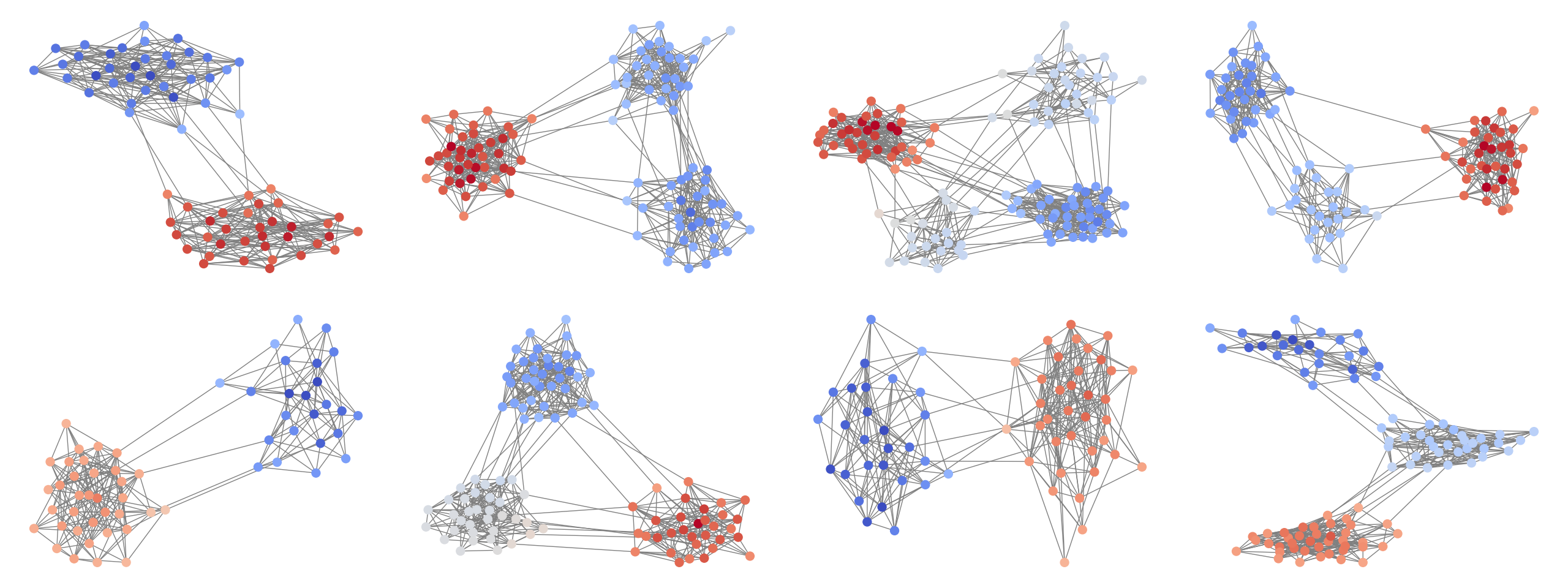}
         \caption{Generated graphs.}
    \end{subfigure}
    \hfill
    \caption{Visualization for SBM dataset.
    }
    \label{fig:vis_sbm}
\end{figure}

\begin{figure}[H]
    \centering
    \begin{subfigure}[b]{0.99\linewidth}
         \centering
         \includegraphics[width=0.9\linewidth]{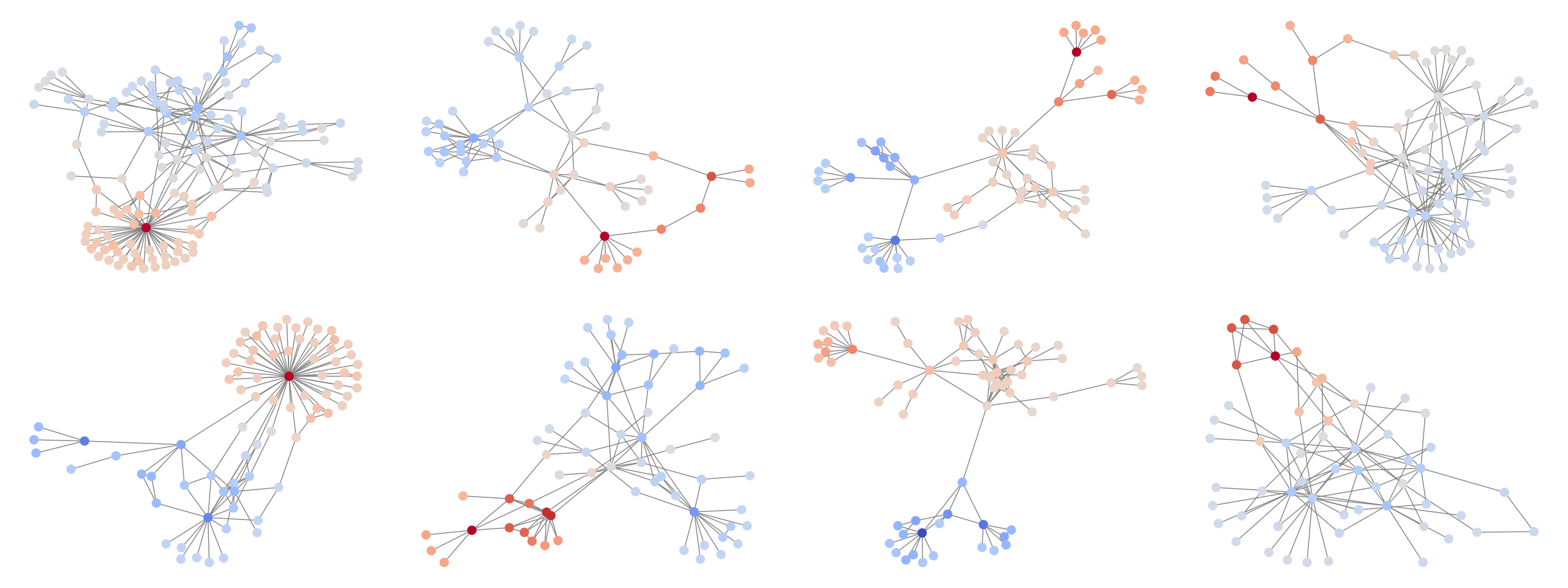}
         \caption{Training graphs.}
    \end{subfigure}
    \hfill
    \begin{subfigure}[b]{0.99\linewidth}
         \centering
         \includegraphics[width=0.9\linewidth]{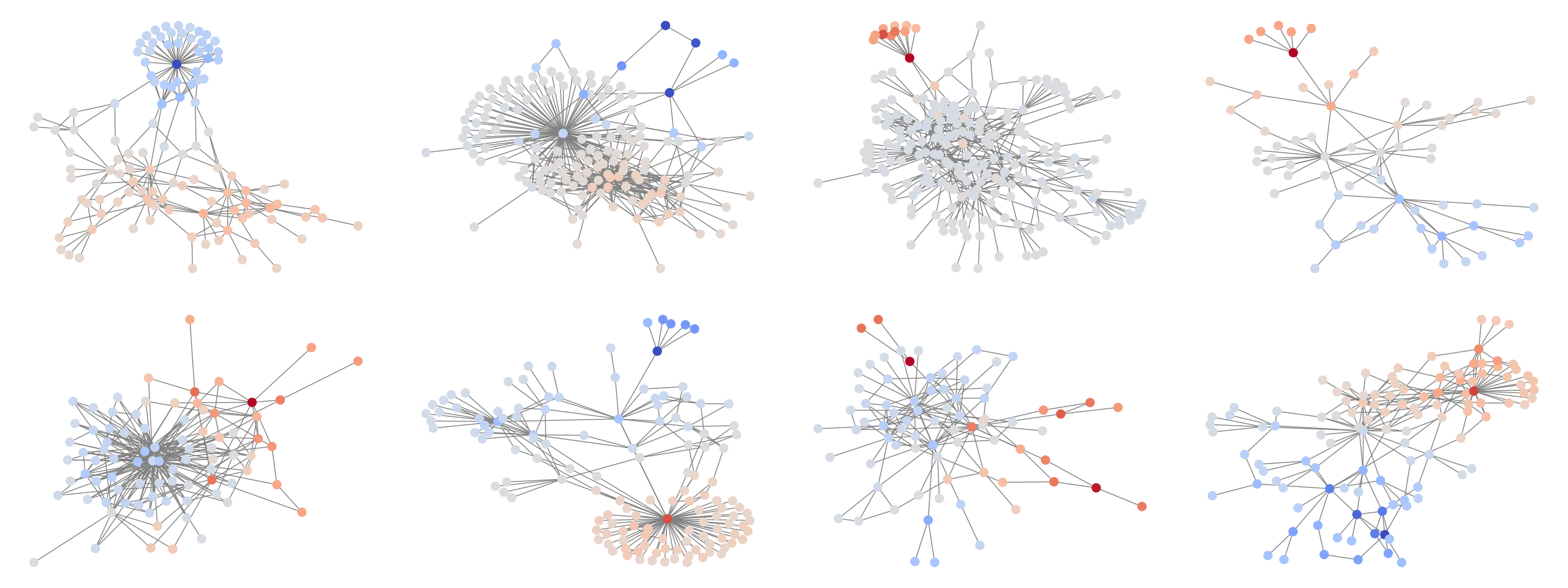}
         \caption{Generated graphs.}
    \end{subfigure}
    \hfill
    \caption{Visualization for Ego dataset.
    }
    \label{fig:vis_ego}
\end{figure}

\begin{figure}[H]
    \centering
    \begin{subfigure}[b]{0.99\linewidth}
         \centering
         \includegraphics[width=0.9\linewidth]{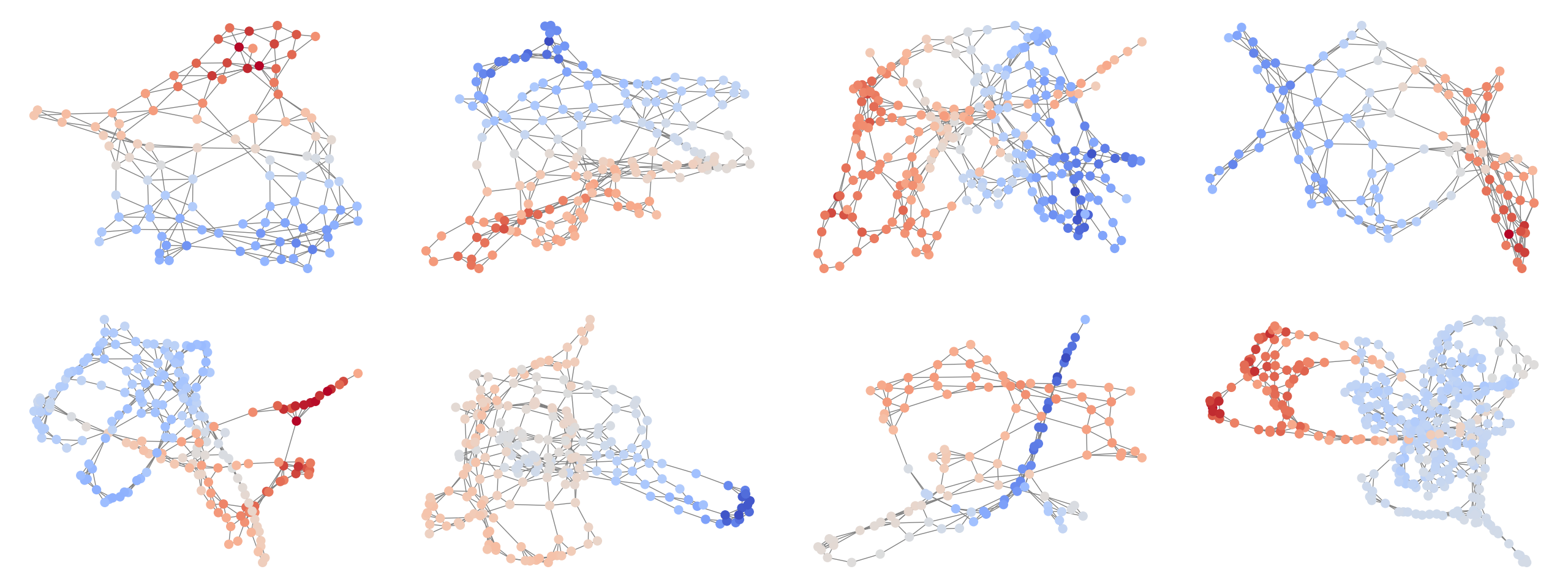}
         \caption{Training graphs.}
    \end{subfigure}
    \hfill
    \begin{subfigure}[b]{0.99\linewidth}
         \centering
         \includegraphics[width=0.9\linewidth]{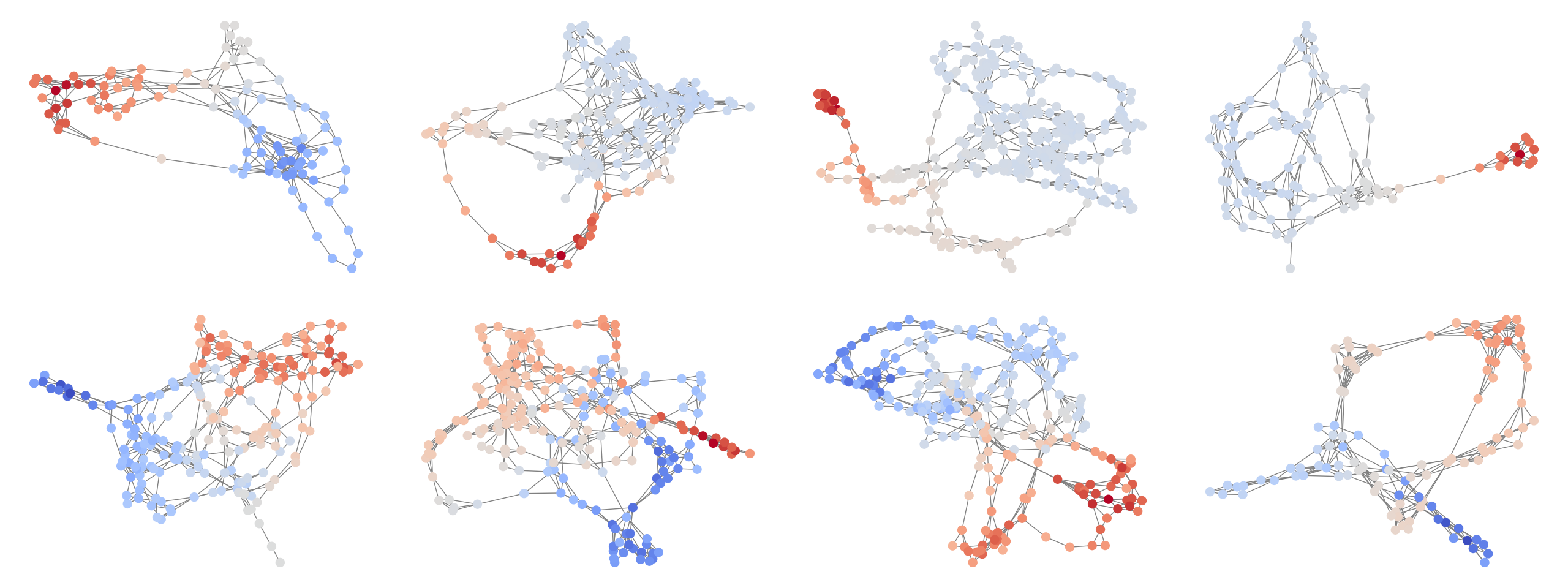}
         \caption{Generated graphs.}
    \end{subfigure}
    \hfill
    \caption{Visualization for Protein dataset.
    }
    \label{fig:vis_protein}
\end{figure}

\end{document}